%% file: 00.main.tex
\pdfoutput=1
\documentclass[11pt]{article}

\usepackage[final]{acl}

\usepackage{times}
\usepackage{latexsym}

\usepackage[T1]{fontenc}
\usepackage[utf8]{inputenc}

\usepackage{microtype}

\usepackage{inconsolata}

\usepackage{graphicx}

\usepackage{bm}
\usepackage{amsmath}
\usepackage{booktabs}
\usepackage{multirow}
\usepackage{xspace}
\usepackage{pifont}
\usepackage{makecell}
\newcommand\our{\textsc{\mbox{MELLE}}}

\usepackage{xcolor}

\usepackage{threeparttable}

\DeclareMathOperator*{\argmax}{arg\,max}

\newcommand{\cmark}{\ding{51}\xspace}
\newcommand{\xmark}{\ding{55}\xspace}
\newcommand{\omark}{\ding{70}\xspace}

\title{Autoregressive Speech Synthesis without Vector Quantization}

\author{
 \textbf{Lingwei Meng\textsuperscript{1*}},
 \textbf{Long Zhou\textsuperscript{2\dag}},
 \textbf{Shujie Liu\textsuperscript{2}},
 \textbf{Sanyuan Chen\textsuperscript{*}},
 \textbf{Bing Han\textsuperscript{2}},
 \textbf{Shujie Hu\textsuperscript{1}},\\
 \textbf{Yanqing Liu\textsuperscript{2}},
 \textbf{Jinyu Li\textsuperscript{2}},
 \textbf{Sheng Zhao\textsuperscript{2}},
 \textbf{Xixin Wu\textsuperscript{1}},
 \textbf{Helen Meng\textsuperscript{1\dag}},
 \textbf{Furu Wei\textsuperscript{2}}
\\
\\
 \textsuperscript{1} The Chinese University of Hong Kong\\
 \textsuperscript{2} Microsoft Corporation\\
 \texttt{\{lmeng,sjhu,wuxx,hmmeng\}@se.cuhk.edu.hk}\\
 \texttt{\{lozhou,shujliu,yanqliu,jinyli,szhao,fuwei\}@microsoft.com}
}

\begin{document}
\maketitle

\begingroup\def\thefootnote{*}\footnotetext{Contribution during an internship at Microsoft Research.}\endgroup
\begingroup\def\thefootnote{\dag}\footnotetext{Corresponding author.}\endgroup

\begin{abstract}
    We present \our{}, a novel continuous-valued token based language modeling approach for text-to-speech synthesis (TTS). \our{} autoregressively generates continuous mel-spectrogram frames directly from text condition, bypassing the need for vector quantization, which is typically designed for audio compression and sacrifices fidelity compared to continuous representations. Specifically, (i) instead of cross-entropy loss, we apply regression loss with a proposed spectrogram flux loss function to model the probability distribution of the continuous-valued tokens; (ii) we have incorporated variational inference into \our{} to facilitate sampling mechanisms, thereby enhancing the output diversity and model robustness. Experiments demonstrate that, compared to the two-stage codec language model VALL-E and its variants, the single-stage \our{} mitigates robustness issues by avoiding the inherent flaws of sampling vector-quantized codes, achieves superior performance across multiple metrics, and, most importantly, offers a more streamlined paradigm. The demos of our work are provided at \url{https://aka.ms/melle}.\footnote{In addition to the model described in this paper, we also trained a \our{} model for \textit{Mandarin} text-to-speech, using the same model configurations and training settings as the English version. Please refer to the page for demos.}
\end{abstract}

\input{01.introduction}

\input{02.related_work}

\input{03.method}

\input{04.exp_setup}

\input{05.results_and_discussion}

\input{06.conclusion}

% \clearpage
% \newpage
\input{07.misc}

\bibliography{custom}

\input{98.appendix}

\end{document}

%% file: 01.introduction.tex
% \newpage
% \setcounter{footnote}{0}

\section{Introduction}
% \vspace{-0.5mm}

% Text-to-speech (TTS) synthesis, which aims to generate high-fidelity speech from given text, has witnessed significant advancements driven by the progress in generative artificial intelligence algorithms \cite{shen2018tacotron2, ren2019fastspeech, li2019transformertts}.
% These developments have empowered TTS systems to generate human-level speech with impressive naturalness and intelligibility.
% However, despite these achievements, there remain significant hurdles in developing zero-shot TTS systems that can accurately capture the nuanced prosody, emotion, and individuality of human speech across diverse speaking styles \cite{wang2023valle, le2024voicebox, ju2024naturalspeech3}.

%Recently, large language models (LLMs) have demonstrated remarkable zero-shot and in-context learning capabilities on natural language processing (NLP) tasks \cite{achiam2023gpt4, touvron2023llama, chen2024valle2}. This has also encouraged the exploration of decoder-only language modeling approaches in other fields. 

\begin{figure}[t]
    \centering
    % \vspace{2mm}
    \includegraphics[width=\columnwidth]{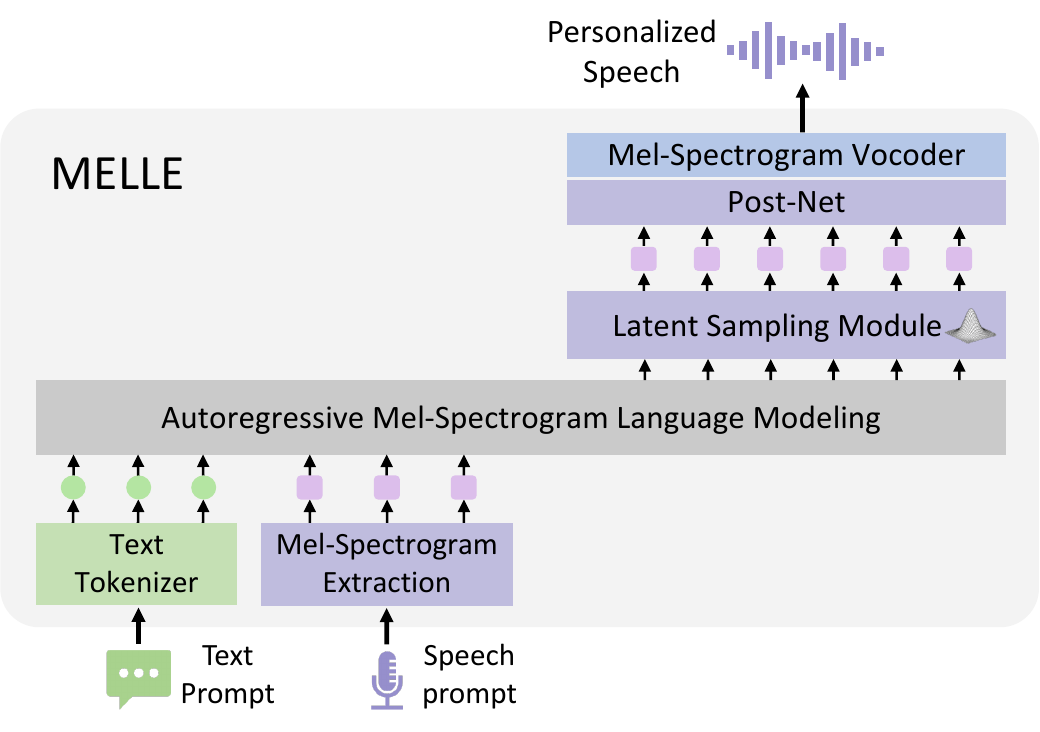}
    % \vspace{3mm}
    \caption{
    Overview of \our{}. 
    Unlike discrete-valued tokens based language modeling, 
    \our{} samples the variational mel-spectrogram conditioned on text and audio prompts using a single-stage decoder-only structure, coupled with the Latent Sampling Module. 
% Unlike discrete-valued tokens based language modeling approaches, MELLE generates the continuous variational mel-spectrogram conditioned on textual and acoustic prompts, using a single-stage decoder-only model as its foundational structures, coupled with the Latent Sampling Module.
    }
    % \vspace{-1.5mm}
    \label{fig:overview}
\end{figure}

The objective of next-token prediction, which involves predicting the next discrete token based on the
previous tokens as a condition, is foundational to the recent progress observed in large language models (LLMs) \cite{brown2020language,OpenAI2023GPT4TR, chen2024ntp}.
%Given an output vector for each token, we can perform the next token prediction by i) taking the output vector for a token and ii) using this to predict the token that comes next in the sequence.
Recently, the success of LLMs in natural language processing (NLP) tasks has encouraged the exploration of autoregressive language modeling approaches in audio synthesis fields. 
Neural codec language models, exemplified by VALL-E \cite{wang2023valle,zhang2023vallex}, reveal the potential of such principle in the zero-shot text-to-speech (TTS) task by leveraging large-scale multi-lingual multi-speaker multi-domain training corpus. 
Unlike traditional TTS systems that rely heavily on complex multi-step pipelines, they utilize a decoder-only structure to predict discrete codec codes, which are vector-quantized tokens encoded from continuous waveforms leveraging neural codec models \cite{zeghidour2021soundstream,dfossez2023encodec}. 

% \begin{figure}[ht!]
%     \centering
%     \includegraphics[width=\columnwidth]{figures/melle_lmeng_motivation.pdf}
%     \caption{Overview of \our{}. Unlike discrete-valued tokens based language models, \our{} samples the continuous variational mel-spectrogram conditioned on textual and acoustic prompts, using a single-stage decoder-only structure coupled with the Latent Sampling Module. 
%     }
%     \label{fig:overview}
% \end{figure}

% VALL-E and its vairants的缺点：1) 采样离散code鲁棒性差，不稳定，容易出现静音；2) codec codes 本身保真度不高；3）two-pass；
Despite achieving impressive naturalness and diversity in synthesized audios, codec language models are plagued by several drawbacks.
First, quantized codec codes, which are typically designed for audio compression, exhibit lower fidelity compared to continuous audio representations if the bit rate is not sufficiently high \cite{puvvada2024discrete, liu2024ardit, bai2024dmel}. Most codec models are trained with mel-spectrogram reconstruction loss, such as SoundStream \cite{zeghidour2021soundstream}, EnCodec \cite{dfossez2023encodec},  and DAC \cite{kumar2023dac}, suggesting that they acquire knowledge from the denser continuous mel-spectrogram space. Some information can be lost after training, even though this information cannot be perceived by the human ear or a specific model.
The similar phenomenon is observed in the field of graphics, where the reconstruction quality of vector-quantized tokens typically lags behind that of their continuous-valued counterparts
\cite{tschannen2023givt,kaiming2024autoregressive}.
Second, neural codec language models suffer from robustness issues stemming from their random sampling strategy, which is inherited from text language models for selecting discrete tokens.
%which requires the empirical assignment of a top-$p$ value to balance the trade-off between unstable outputs and infinite repetitions. 
This issue is more pronounced with acoustic tokens compared to textual ones due to the greater similarity among consecutive codec codes, which can result in  extended stretches of silence or persistent noise \cite{song2024ellav}.
%Compared to textual tokens, this drawback is exacerbated by the high homogeneity between adjacent acoustic tokens, resulting in endless silence or noise \cite{wang2023valle, song2024ellav}. 
Third, neural codec language models typically necessitate a complicated two-pass decoding process, involving an autoregressive (AR) model for generating coarse primary audio tokens, followed by a non-autoregressive (NAR) model to iteratively predict the remaining multi-codebook codes for refinement. 
This multi-step process compromises inference efficiency, leading to increased computational and storage demands. 

% Some subsequent works based on VALL-E attempts to alleviate these issues. Most of their efforts focus on introducing additional supervision and constraints to drive the model to grasp robustness through elaborately designed monotonic alignment \cite{song2024ellav,du2024vallt, han2024valler}. Another work \cite{xin2024ralle} autoregressively predicts pitch and duration tokens before codec codes, which also explicitly increases the length of the generated sequences. However, the improvements are limited because these methods do not circumvent the natural drawbacks of randomly sampling discrete acoustic tokens, and they increase the model's complexity and inference.
% VALL-E 2 \cite{chen2024valle2} introduces a repetition aware sampling to stabilize the decoding and alleviate infinite loop issue. While this requires manually setting additional sampling parameters which can be domain-specific. 
% A recent work \cite{kim2024clamtts} predicts latent representations of a neural codec model instead of discrete codes, achieving better results. However, the performance is still constrained by the highly compressed nature of the codec.

% Spectron \cite{nachmani2024spectron}

% many works on robusteness, mono alig, with additional supervision
% also poses robustness issues such as repetitions and omissions. 

To address the limitations associated with discrete-token-based codec language models, we are rethinking the potential of continuous representations and aim to determine whether continuous-valued tokens can supplant discrete-valued tokens within the paradigm of autoregressive speech synthesis models.
The successful implementation of the autoregressive model without vector quantization faces two key challenges:
\textit{\textbf{(i) How to set training objectives for continuous representation?}} 
The continuous space significantly differs from that of vector-quantized tokens, for which autoregressive language models typically adopt a next-token prediction objective, with cross-entropy loss to measure the discrepancy between the predicted probabilities and the targets. 
\textit{\textbf{(ii) How to enable sampling mechanism in continuous space?}} The sampling strategy is a critical component in both text generation and speech synthesis systems, as it introduces diversity into the output and enhances their generalization ability. However, continuous-valued token based models can not employ top-p random sampling method used in discrete codec language models.
%Discrete codec language models adopt top-p random sampling with a softmax function to avoid the collapsing of outputs that can be caused by a greedy search.

In this work, we propose \our{}, a robust single-pass zero-shot TTS model that autoregressively predicts continuous mel-spectrogram\footnote{We leave the exploration of other continuous representations, such as VAE latent states, for future research endeavors.} frames based on previous tokens. 
In response to the aforementioned challenges, we first substitute cross-entropy loss with regression loss and introduce a spectrogram flux loss to promote variation in the prediction and eliminate repetition issues.
Second, we design a latent sampling module, derived from variational inference, functioning as a sequence sampling strategy thereby enhancing the diversity of the generated audios. 
As an option, by adjusting the reduction factor, \our{} can predict multiple frames per step and accelerate inference, thereby further alleviating robustness issues associated with long-sequence modeling and maintaining satisfactory performance.

We conducted evaluations of the proposed \our{} on both the large-scale 50K-hour Libriheavy \cite{kang2024libriheavy} training dataset and the relatively small 960-hour LibriSpeech \cite{panayotov2015librispeech} training dataset. 
We use LibriSpeech test-clean set for zero-shot TTS evaluation.
Experimental results demonstrate that the proposed \our{} is on par with VALL-E 2 \cite{chen2024valle2} in objective metrics, and surpasses VALL-E 2 in subjective metrics. It also outperforms previous neural codec language models, including VALL-E and its other variants, achieving superior performance across multiple metrics that reflect naturalness, robustness, similarity, and inference efficiency. 
Specifically, \our{} surpasses the ground truth audios in WER (1.47\% vs. 1.61\%), achieving a 47.9\% relative reduction in WER compared to VALL-E and an 8.1\% reduction compared to VALL-E 2 on the continuation inference task for zero-shot TTS. 
% Specifically, \our{} surpasses the ground truth audios in WER (1.47\% vs. 1.61\%) on both continuation and cross-sentence zero-shot TTS tasks. 
For subjective evaluations, \our{} is more favorably received by human listeners than previous models, achieving comparable performance to the ground truth in terms of MOS (4.20 vs. 4.29) and CMOS (\mbox{-0.032} vs. ground truth), and an even higher SMOS (4.40 vs. 3.94) than the ground-truth speech.

%% file: 02.related_work.tex
\section{Related Work}

% \vspace{-0.5mm}

% \begin{figure}[t]
%     \centering
%     \includegraphics[width=\columnwidth]{figures/melle_lmeng_overview1.pdf}
%     \caption{Overview of \our{}. Unlike discrete-valued tokens based language models, \our{} samples the continuous variational mel-spectrogram conditioned on textual and acoustic prompts, using a single-stage decoder-only structure coupled with the Latent Sampling Module. 
%     }
%     \label{fig:overview}
% \end{figure}

% \subsection{Traditional TTS}
% Traditional speech synthesis methods can be categorized concatenative systems, parametric systems, and end-to-end neural systems. Concatenative TTS systems deconstruct original audio waves into smaller segments and then reassembles them using algorithms like Viterbi, followed by signal processing techniques, to create new audio waves \cite{hunt1996unit,black1997automatically}.
% Parametric TTS systems convert speech waves into spectrograms, utilizing acoustic parameters such as fundamental frequency and duration to synthesize new audio outputs \cite{zen2013statistical,tokuda2013speech}.
\paragraph{End-to-End TTS}
% With the rapid development of neural networks, end-to-end neural TTS systems are proposed to simplify previous speech synthesis pipeline via a single neural network.
% eliminating the need for the production of these linguistic and acoustic features \cite{wang2017tacotron,li2019transformertts,ren2019fastspeech}.
End-to-end neural TTS models are proposed to simplify the previous pipeline by using a single neural network. These models typically generate mel-spectrograms directly from text and then recover the audio from the mel-spectrograms using a vocoder. TransformerTTS \cite{li2019transformertts} employs Transformer encoder-decoder network as the backbone to replace RNN structures in Tacotron \cite{wang2017tacotron}. FastSpeech \cite{ren2019fastspeech} further improve the speech quality and decoding efficiency using the non-autoregressive generation model with a duration module. These models are trained on small-scale, clean, single- or few-speaker dataset. Our \our{} leverages the well-established mel-spectrogram as the target representation, however, it differs significantly in two key aspects: (1) We adopt decoder-only network as foundational structure with improved methods, such as variational inference and spectrogram flux loss, (2) \our{} is capable of zero-shot TTS via language modeling training on large-scale data.
% , like Libriheavy \cite{kang2024libriheavy}.

% \begin{figure}[t]
%     \centering
%     \vspace{-1mm}
%     \includegraphics[width=\columnwidth]{figures/melle_lmeng_overview3.pdf}
%     \vspace{-5mm}
%     \caption{Overview of \our{}. Unlike discrete-valued tokens based language models, \our{} samples the continuous variational mel-spectrogram conditioned on textual and acoustic prompts, using a single-stage decoder-only structure coupled with the Latent Sampling Module. 
%     }
%     % \vspace{-5mm}
%     \label{fig:overview}
% \end{figure}

% \begin{figure*}[t]
%     \centering
%     \vspace{-3mm}
%     \includegraphics[width=0.8\textwidth]{figures/melle_lmeng_modules2.pdf}
%     \vspace{-1mm}
%     \caption{
%     The Latent Sampling Module (left), Stop Prediction Layer (mid), and Post-Net (right). 
%     }
%     \label{fig:modules}
% % \vspace{-2mm}
% \end{figure*}

% \subsection{Zero-Shot TTS}
% \vspace{-0.5mm}
\paragraph{Zero-Shot TTS}
Motivated by the in-context learning abilities of LLMs on NLP tasks, various studies are proposed to address zero-shot TTS through a language modeling approach.
VALL-E \cite{wang2023valle, zhang2023vallex} first utilizes codec codes as intermediate representation, then uses a codec decoder to reconstruct the audio.
% VALL-E necessitates two-stage modeling, with an autoregressive model for generating coarse audio tokens, followed by a non-autoregressive model to iteratively predict multi-codebook codes for refinement. 
% VALL-E X \cite{zhang2023vallex} extends VALL-E into multi-lingual scenario to support zero-shot cross-lingual speech synthesis and speech-to-speech translation.
Mega-TTS \cite{jiang2023mega} proposes to disentangle the multiple attributes in speech, such as content, timbre, prosody, and phase,
then model them with a language model.
ELLA-V \cite{song2024ellav}, RALL-E \cite{xin2024ralle}, and VALL-E R \cite{han2024valler} aims to improve robustness of VALL-E via additional fine-grained speech-text alignments. 
BASE TTS \cite{lajszczak2024base} employs discrete tokens derived from WavLM \cite{chen2022wavlm} and scales the language model to larger size and training data. 
Parallel to our work, VALL-E 2 \cite{chen2024valle2} shares the same architecture as VALL-E but employs a repetition-aware sampling strategy that promotes more deliberate sampling choices. 
% Seed-TTS \cite{anastassiou2024seed} replaces NAR model with a diffusion model, which generates continuous speech representations according generated speech tokens from AR stage. 
Rather than using an NAR model to generate residual discrete codes, some works employ diffusion or flow-matching as the second stage to reconstruct mel-spectrograms or other continuous representations, such as TorToise-TTS \cite{betker2023tortoise}, CosyVoice \cite{du2024cosyvoice}, and SEED-TTS \cite{anastassiou2024seed}. They indicate that operations in continuous spaces yield improved performance. However, they still necessitate two-stage modeling, unlike MELLE, which requires only single-stage modeling.

Other studies have investigated fully non-autoregressive approaches. SoundStorm \cite{borsos2023soundstorm} adapts a parallel, confidence-based decoding scheme for generating codec codes.
StyleTTS 2 \cite{li2024styletts} and NaturalSpeech 3 \cite{ju2024naturalspeech3} use diffusion model to achieve better TTS synthesis.
Voicebox \cite{le2024voicebox} and Audiobox \cite{vyas2023audiobox} employ flow-matching based models for transcript-guided speech generation.
Recently, E2 TTS \cite{eskimez2024e2} presents a TTS systems consisting of flow-matching-based mel-spectrogram generator trained with the audio infilling task. 
Different from previous works, \our{} is a continuous-valued token based autoregressive language model with variational inference for text-to-speech synthesis, striving to achieve higher fidelity and naturalness.

%% file: 03.method.tex
\begin{figure*}[t]
    \centering
    \vspace{-2mm}
    \includegraphics[width=1\textwidth]{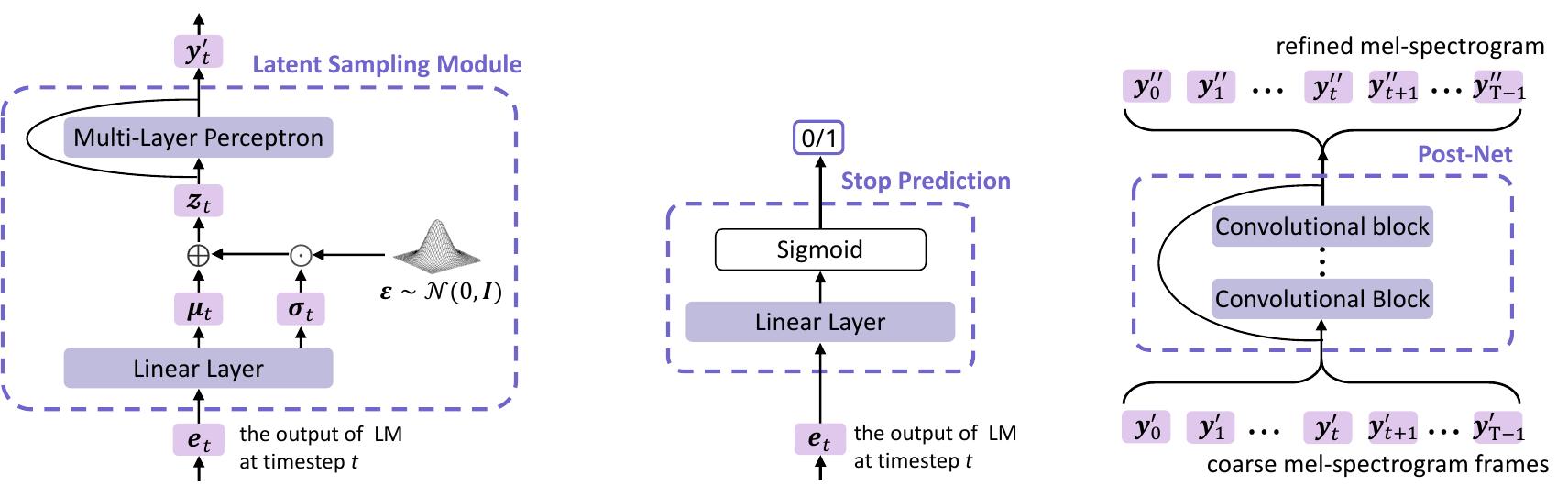}
    % \vspace{-6mm}
    \caption{
    The Latent Sampling Module (left), Stop Prediction Layer (mid), and Post-Net (right). 
    }
    \label{fig:modules}
    % \vspace{-4mm}
\end{figure*}

% \vspace{-1mm}
\section{\our{}}
% \vspace{-1mm}
\subsection{Problem Formulation}
% \vspace{-0.5mm}
This study regards TTS as an autoregressive mel-spectrogram language modeling task.
% Instead of predicting highly-compressed discrete codec codes like VALL-E, \our{} directly predicts continuous mel-spectrograms which are then converted to waveforms with an off-the-shelf vocoder, striving to achieve higher fidelity and naturalness.
Given the byte-pair-encoded (BPE) text content $\bm{x} = [x_0, x_1, \ldots, x_{L-1}]$ of an audio sample, 
%and the speech prompt in mel-spectrogram format $\Tilde{\bm{y}}$, \
\our{} is optimized to predict the mel-spectrogram $\bm{y} = [\bm{y}_0, \bm{y}_1, \ldots, \bm{y}_{T-1}]$ extracted from the audio. Specifically, at each autoregressive step, \our{} is expected to predict the next mel-spectrogram frame $\bm{y}_{t}$ conditioned on the text prompt $\bm{x}$ and the previous mel-spectrograms $\bm{y}_{<t}$, which is equivalent to maximizing the following distribution:
% \vspace{-1mm}
\begin{align}
   p(\bm{y} \mid \bm{x}; \bm{\theta}) &= \prod_{t=0}^{T-1} p(\bm{y}_t \mid \bm{y}_{<t}, \bm{x}; \bm{\theta})
\end{align}
% \vspace{-3mm}

\noindent where $\bm{y}_{<t}$ denotes $[\bm{y}_0, \bm{y}_1,..., \bm{y}_{t-1}]$ and $\bm{\theta}$ represents the parameters of \our{}. 

Inspired by previous neural TTS models \cite{li2019transformertts}, we introduce a reduction factor $r$ to control the number of mel-spectrogram frames predicted at each decoding step, providing a balance between computational efficiency and generation quality. Formally, the original mel-spectrogram sequences $\bm{y}$ will be partitioned into 
% $y^r = [y_{0:r}, y_{r:2r}, ..., y_{(T-r):T}]$ 
$\bm{y}^r = [\bm{y}_{0:r}, \bm{y}_{r:2r}, ..., \bm{y}_{(T-r):T}]$ 
with a factor $r$, and the likelihood function can be expressed as
% \vspace{-1mm}
\begin{align}
   \hspace{-2.1mm} p(\bm{y} \mid \bm{x}; \bm{\theta}) = \prod_{t=0}^{T/r-1} p(\bm{y}_{t\cdot r:(t+1)\cdot r} |\ \bm{y}_{<t\cdot r}, \bm{x}; \bm{\theta})
\end{align} 
% \vspace{-3mm}

During inference, \our{} executes zero-shot TTS via prompting. Given the text content $\bm{x}$ for synthesis, the text transcript $\Tilde{\bm{x}}$ and mel-spectrogram $\Tilde{\bm{y}}$ of speech prompt, the model is designed to generate the target mel-spectrogram $\bm{y}$ corresponding to $\bm{x}$ while preserving the characteristics of the speaker in prompt, 
%by concatenating $\Tilde{\bm{x}}$, $\bm{x}$, and $\Tilde{\bm{y}}$ as model inputs, 
with maximum likelihood probability as $\argmax_{\bm{y}} p(\bm{y}_{t\cdot r:(t+1)\cdot r} \mid [\Tilde{\bm{x}}; \bm{x}; \Tilde{\bm{y}}; \bm{y}_{<t\cdot r}]; \bm{\theta})$ at each time step, and it backs to standard mode if $r=1$.

% \vspace{-0.5mm}
\subsection{\our{} Architecture}
% transformer decoder
% prenet
% sampling module (projection and sampling)
% postnet   (CNN)
% stop prediction
% reduction factor

As illustrated in Figure \ref{fig:overview}, \our{} comprises the following main components: pre-nets that respectively convert text into sub-word tokens and extract mel-spectrograms from speech, before projecting them to the model dimension; an Transformer decoder that serves as the language model; a latent sampling module that samples latent embedding from a predicted distribution, and then projects it back to the spectrogram space; a stop prediction layer to determine the end of the generation and a convolutional post-net for spectrogram refinement. Finally, a vocoder is used to recover the speech from generated mel-spectrogram. 

% As illustrated in Figure \ref{fig:overview}, \our{} comprises the following main components: an autoregressive (AR) transformer decoder serving as the language model, a sampling module to sample latent embedding from a Gaussian distribution and project them back to the spectrogram space, a stop prediction layer, and a convolutional block as the post-net for spectrogram refinement following \cite{shen2018tacotron2,li2019transformertts}. 

% As illustrated in Figure \ref{fig:overview}, within \our{}, the text will be converted into sub-word tokens by BPE and 

Unlike neural codec language models that iteratively predict multi-layer codec codes, we do not require an additional non-autoregressive (NAR) model thanks to the completeness of the mel-spectrogram. This simplification significantly improve computational and storage efficiency. 
Moreover, by adjusting the reduction factor, \our{} can generate multiple mel-spectrogram frames at one step, further enhancing efficiency while still maintaining superior performance.

% \noindent \textbf{Autoregressive transformer decoder.} 
% \vspace{-1mm}
\subsubsection{Autoregressive Language Model}
% \vspace{-0.5mm}

We employ an Transformer decoder as the language model (LM) to autoregressively generates acoustic continuous tokens based on the textual and acoustic prompts. Specifically, input text tokens $\bm{x}$, with an appended \textless EOS\textgreater\ token, are first converted into embeddings by the text embedding layer based on their indices. 
Simultaneously, we employ a multi-layer perceptron, named pre-net, to project the mel-spectrogram $\bm{y}$  to the language model dimension. %$<\Tilde{\bm{y}}, \bm{y}_{<t}>$
% Simultaneously, the mel-spectrogram of the speech prompt is projected to the LM's dimension via acoustic projection layers. 
The LM, consisting of blocks of multi-head attention and feed-forward layers, takes the concatenation of text and acoustic embeddings as input to model the dependency between semantic and acoustic information. The output of the LM $\bm{e}_t$ at time step $t$ is subsequently processed by the following modules of \our{} to synthesize the next-frame output, which is detailed below.

% The output of the LM is projected to a latent embedding $\bm{z}_t$ at time step $t$, which is an intermediate product with the same dimensions as the mel-spectrogram.
% $\bm{z}_t$ is subsequently processed by the following modules of \our{} to synthesize the next-frame output.

% \vspace{-1mm}
\subsubsection{Latent Sampling Module}
% \vspace{-0.5mm}
The sampling strategy is a critical part in TTS systems, as it not only introduces diversity in the output, but also enhances generalization ability. 
For example, Tacotron \cite{wang2017tacotron} enable dropout in their pre-net during inference to introduce variation;
Codec language models \cite{wang2023valle} adopt the top-p random sampling to avoid the collapse outputs leading by greedy search; Diffusion-based \cite{ju2024naturalspeech3} and flow-matching-based methods \cite{le2024voicebox} restore speech representations from the sampling of a simpler distribution. 
%In this study, we draw inspiration from variational auto-encoder (VAEs) \cite{kingma2014vae, blei2017vi_review}. Instead of borrowing an external VAE model, 

In this study, inspired by variational autoencoder (VAE) \cite{kingma2014vae}, we integrate a novel latent sampling module 
% to explicitly model a distribution
within \our{}, aimed at enhancing both expressive diversity and robustness, as shown in Figure \ref{fig:modules} (left). Based on the LM output $\bm{e}_t$, this module predicts a distribution, from which a latent embedding $\bm{z}_t$ is sampled. %This latent embedding is then mapped to the mel-spectrogram space.

Specifically, we assume that $\bm{z}_t$ follows a multivariate Gaussian distribution where each dimension is independent. As depicted in Figure \ref{fig:modules}, a linear layer ($\mathbf{W} [\cdot] + \mathbf{b}$) predicts a mean vector $\boldsymbol{\mu}_t$ and a log-magnitude variance vector $\log\bm{\sigma}_t^2$ of the Gaussian distribution based on $\bm{e}_t$.
Leveraging the reparameterization technique, a $\bm{z}_t$ is sampled as
% \vspace{-1mm}
\begin{align}
\bm{z}_t = \bm{\mu}_t + \bm{\sigma}_t \odot \bm{\epsilon}
\end{align}
% \vspace{-5.5mm}

\noindent where $\bm{\epsilon} \sim \mathcal{N}(0, \bm{I}),\ \ [\bm{\mu}_t, \log \bm{\sigma}_t^2] = \mathbf{W} \bm{e}_t + \mathbf{b}$. Then, the probability density function is defined as
% \vspace{-5mm}
\begin{align}
\label{eq:gaussian}
&p_{\bm{\theta}}(\bm{z}_t \mid \bm{e}_t) = \mathcal{N}(\bm{z}_t \mid \bm{\mu}_t, \mathrm{diag}(\bm{\sigma}_t^2)) 
% \\&\text{where } [\bm{\mu}_t, \log \bm{\sigma}_t] = \mathbf{W} \bm{e}_t + \mathbf{b}
\end{align}
% \vspace{-5mm}

\noindent Note that it is differentiable with the reparameterization technique. Next, the latent variable $\bm{z}_t$ is passed through a multi-layer perceptron with residual connections, mapping it to the mel-spectrogram space as $\bm{y}'_{t}$, where $t=0, 1, ..., T-1$.

%%%%%%%%%%%%%%%%%%%%
% \paragraph{Post-Net}
% Following previous neural TTS models \cite{wang2017tacotron, shen2018tacotron2, li2019transformertts}, we employ a multi-layer convolutional block to produce a residual that is added to $\bm{y}'$, resulting in the refined mel-spectrogram $\bm{y}''$. Since \our{} directly predicts continuous mel-spectrograms rather than discrete tokens, it cannot generate an \textless EOS\textgreater\ token to indicate the end of generation. Instead, we use a simple linear layer as a binary classifier, taking $\bm{e}_t$ to determine if the generation should conclude.
%%%%%%%%%%%%%%%%%%%%

% \vspace{-0.5mm}
\subsubsection{Stop Prediction Layer and Post-Net}
We use a linear layer as a binary classifier, taking $\bm{e}_t$ to determine if the generation should conclude, as depicted in Figure \ref{fig:modules} (mid).
Following previous neural TTS models \cite{li2019transformertts}, we employ multiple convolutional blocks as the post-net to produce a residual that is added to $\bm{y}'=\{\bm{y}'_{0}, \bm{y}'_{1}, ..., \bm{y}'_{T-1}\}$, resulting in the refined mel-spectrogram $\bm{y}''=\{\bm{y}''_{0}, \bm{y}''_{1}, ..., \bm{y}''_{T-1}\}$, as shown in Figure \ref{fig:modules} (right). 
During training, the model is trained using teacher-forcing; while during inference, post-net processes $\bm{y}'$ after the AR generation concludes.

% \vspace{-1mm}
\subsection{Training Objective}
% L1 and L2
% KL loss
% Spectrogram flux loss
% The three losses work synergistically to enhance the overall performance of the model.
% \noindent \textbf{Spectrogram flux loss.}

The training process of \our{} is efficient and straightforward, due to the absence of \mbox{VALL-E}'s complex hierarchical structure. As illustrated in Figure \ref{fig:overview}, a single end-to-end autoregressive model is optimized during training in teacher-forcing manner using four loss functions: (1) a regression loss; (2) a Kullback-Leibler (KL) divergence loss; (3) a novel spectrogram flux loss; and (4) a binary cross entropy (BCE) loss for stop prediction. They work collaboratively to enhance overall performance:
% \vspace{-1mm}
\begin{align}
\mathcal{L} = \mathcal{L}_{\text{reg}} + \lambda \mathcal{L}_{\text{KL}} + \beta\mathcal{L}_{\text{flux}}  + \gamma \mathcal{L}_{\text{stop}}
\end{align}
% \vspace{-5mm}

\paragraph{Regression Loss}
The regression loss is a fundamental component of the training objective, ensuring the accurate prediction of mel-spectrogram frames. The regression loss, \(\mathcal{L}_{\text{reg}}\), is composed of a combination of L1 and L2 losses, applied to both intermediate prediction $\bm{y}'$ and final prediction $\bm{y}''$ of the mel-spectrogram. It is defined as follows:
% \vspace{-2mm}
\begin{align}
\mathcal{L}_{\text{reg}}(\bm{y}, \bm{y}', \bm{y}'') =  \ &\|\bm{y} - \bm{y}'\|_1 + \|\bm{y} - \bm{y}'\|_2^2 \notag \\
    + & \|\bm{y} - \bm{y}''\|_1 + \|\bm{y} - \bm{y}''\|_2^2
\end{align}
% \vspace{-6mm}

\noindent where $\bm{y}$ is the ground-truth spectrogram target.

% \vspace{-1mm}
\paragraph{KL Divergence Loss}
%We assume that the latent embedding \(\bm{z}_t\) conforms to a multivariate Gaussian distribution. By sampling from this distribution, we aim to enhance the diversity and stability of our generated outputs. To formalize this assumption in our model training, we introduce the Kullback-Leibler (KL) divergence loss based on the concept of variational inference \cite{kingma2014vae, blei2017vi_review}.
We introduce a Kullback-Leibler (KL) divergence loss based on the concept of variational inference \cite{kingma2014vae}, to enhance the diversity and stability of \our{}.
The KL divergence measures the difference between the predicted latent distribution \mbox{\( p_{\bm{\theta}}(\bm{z}_t \mid \bm{e}_t) \)} and a simpler distribution \( p(\bm{z}_t) \). Unlike \citet{kingma2014vae}, which selects \( p(\bm{z}_t) \) as a standard normal distribution, we let \(\bm{z}_t\) possess the same dimensionality as the mel-spectrogram and define \( p(\bm{z}_t) \) as \(\mathcal{N}(\bm{y}_t, \bm{I})\). 
This can be seen as a shortcut on the optimization path thus accelerates the model's learning.
Combining equation (\ref{eq:gaussian})
% \vspace{-2mm}
\begin{align}
&\mathcal{L}_\text{KL}(\bm{y}, \bm{z}) = \sum_{t=0}^{T-1} D_{\text{KL}}(p_{\bm{\theta}}(\bm{z}_t \mid \bm{e}_t) \parallel p(\bm{z}_t)) \notag \\
&\hspace{-1.5mm}=\frac{1}{2} \sum_{t=0}^{T-1} ( \|\bm{\sigma}_t\|^2_2 + \|\bm{\mu}_t - \bm{y}_t\|^2_2 - d - \sum_{i=1}^d \log \bm{\sigma}_t^2[i] )
\end{align}
% \vspace{-6mm}

\noindent where \( d \) is the dimensionality of the feature space. The detailed derivation is provided in Appendix \ref{appendix:derivation}. By integrating the KL divergence loss, \our{} achieves a balance between synthesis quality and latent space regularization, ultimately enhancing the expressive diversity and robustness of the generated mel-spectrograms.

% the KL loss component helps the model maintain a structured latent space, ensuring that the latent variables adhere closely to the prior distribution. By integrating the KL divergence into the loss function, \our{} achieves a balance between reconstruction quality and latent space regularization, ultimately contributing to enhanced expressive diversity and robustness of the generated mel-spectrograms.

\paragraph{The Spectrogram Flux Loss}
To encourage dynamic variation in the generated frames, a novel spectrogram flux loss is proposed as a regularization term that penalizes low variability between consecutive frames and promotes changes:
% \vspace{-2mm}
\begin{align}
\mathcal{L}_{\text{flux}}(\bm{y}, \bm{\mu}) = - \sum_{t=1}^{T-1} \|\bm{\mu}_t - \bm{y}_{t-1}\|_1
\end{align}
% \vspace{-3mm}
\noindent where the L1 norm is employed to measure the difference between the predicted Gaussian mean vector $\bm{\mu}_t$ and the previous ground truth frame $\bm{y}_{t-1}$. By summing the negative values of the differences, the loss rewards variations in the generated frames and discourages overly static frames, which can lead to repetition or prolonged silence in synthesized audio. 
By penalizing flat predictions, the model is incentivized to produce more diverse and dynamic spectrograms, thereby preventing monotonic and unnatural speech.
%By integrating this loss into the overall training objective, we found that the model significantly reduces the generation of overly static frames, which can lead to repetition or endless silence in synthesized audio. 
% \vspace{-1mm}

\paragraph{Stop Prediction Loss}
We use a linear layer to project LM output $\bm{e}_t$ to a logit and calculate the BCE loss, $\mathcal{L}_{\text{stop}}$, for stop prediction, similar to SpeechT5 \cite{ao2021speecht5}. 
Considering each utterance has only one positive frame indicating "stop," the positive and negative frames are extremely imbalanced. To address this, we assign a larger weight (100) to the positive frames in the BCE loss.
% \begin{align}
% \mathcal{L}_{stop}(\bm{s}, \bm{s}') = - \sum_{t=1}^{T-1} \left[ 100 \cdot s_t \log(s'_t) + (1 - s_t) \log(1 - s'_t) \right]
% \end{align}

\paragraph{Inference: In-Context Learning}
% \vspace{-1mm}
During inference, we perform zero-shot TTS by autoregressively predicting mel-spectrogram. 
Given the text content $\bm{x}$ and a piece of speech prompt (with text transcription $\Tilde{\bm{x}}$ and mel-spectrogram $\Tilde{\bm{y}}$),
at each time step \( t \), \our{} generates the next-frame \( \bm{y}_t' \) from a latent embedding \( \bm{z}_t \), which is sampled from a distribution conditioned on the concatenation of $\Tilde{\bm{x}}$, $\bm{x}$, $\Tilde{\bm{y}}$, and $\bm{y}_{<t}$.
% the textual prompt \( \bm{x} \) and previous speech frames \( \left< \Tilde{\bm{y}}, \bm{y}_{<t} \right> \). 
After the AR generation process concludes, the coarse mel-spectrogram \( \bm{y}' \) passes through the post-net to obtain the refined spectrogram \( \bm{y}'' \), which is then converted to speech audio using an off-the-shelf vocoder.
If the reduction factor $r$ is set, the input and predicted mel-spectrograms will be grouped by $r$. 

%Unlike neural codec language models, which require multi-stage iterative predictions on multi-codebook codes, \our{} completes the speech synthesis in a single pass. Furthermore, unlike neural codec language models that necessitate manually setting sampling parameters, \our{} automatically samples from learned distributions that are unique to each input. 
Unlike codec language models (e.g., VALL-E) that rely on multi-stage iterative predictions across multi-layer codes and require manual configuration of sampling parameters, \our{} accomplishes speech synthesis in a single forward pass and automatically samples from learned distributions that are unique to each input. 
This automated approach ensures adaptive and consistent sampling, reduces human effort, and makes the method domain-independent.
With the strong in-context learning capability from LM, \our{} is capable of generating high-fidelity, natural-sounding speech for unseen speakers without fine-tuning.

%By leveraging the proposed methods, \our{} is capable of generating high-fidelity, natural-sounding speech for unseen speakers without fine-tuning. This demonstrates its in-context learning capability stemming from its language model design. 

%% file: 04.exp_setup.tex
% \vspace{-1mm}
\section{Experimental Setup}
% \vspace{-1mm}
\subsection{Training Datasets}
% \vspace{-0.5mm}
We trained \our{} on the Libriheavy \cite{kang2024libriheavy} corpus, 
% an annotated extension of the LibriLight \cite{kahn2020librilight} corpus. Libriheavy 
which contains approximately 50K hours of speech from 6,736 speakers, sourced from English audiobooks.
% that are part of the LibriVox project. 
We use byte-pair encoding (BPE) for text tokenization with a vocabulary size of 4K.
For audios, we perform voice activity detection to remove abnormal silences and facilitate training. The 80-dimensional log-magnitude mel-spectrograms are extracted at 62.5 Hz with a window length of 1,024 and a hop length of 256, from waveforms resampled at 16 kHz.
% The detailed extraction protocol of mel-spectrogram can be found in \red{Appendix \ref{appendix:mel}}. 

% Additionally, to facilitate comparison with research conducted under constrained resources \cite{song2024ellav, han2024valler}, we trained a constrained version of \our{} on LibriSpeech \cite{panayotov2015librispeech}, which contains 960 hours of data from 1,251 speakers. We use phoneme text tokens for this version. This limited version, denoted as \mbox{\our{}-\textit{limited}}, aims to verify the effectiveness of our model when trained on smaller dataset.

Additionally, to verify the effectiveness of our method under constrained resources, we trained a limited version of our model, denoted as \our{}-\textit{limited}, on LibriSpeech \cite{panayotov2015librispeech}, which contains 960-hour data from 1,251 speakers. We use phoneme text tokens for this version.

%\subsection{Model Configurations and Training Settings}
% \vspace{-1mm}
\subsection{Experimental Settings}
% \vspace{-0.5mm}

% \paragraph{Model Configurations} 
The LM of \our{} contains 12 Transformer blocks, each with 16 attention heads, an embedding dimension of 1,024, a feed-forward network dimension of 4,096, and a dropout rate of 0.1. The input mel-spectrograms are projected to the LM dimension using a 3-layer perceptron with a 0.5 dropout rate enabled during both training and inference, following Tacotron. Within the latent sampling module, the sampled \(\bm{z}_t\) passes through a 3-layer perceptron to produce a residual, which is then added to itself to generate \(\bm{y}'_t\). The post-net, consisting of 5 convolutional blocks with a kernel size of 5 and 256 intermediate channels, takes \(\bm{y}'\) to generate the refined \(\bm{y}''\).
Throughout this study, 
we utilize an open-source HiFi-GAN vocoder\footnote{The pre-trained vocoder can be found in \url{https://huggingface.co/mechanicalsea/speecht5-tts}} \cite{kong2020hifigan}, trained on LibriTTS, to reconstruct audios from mel-spectrograms.

The training hyper-parameters and details of \our{} can be found in Appendix \ref{appendix:train}.
% \paragraph{Training Details} \our{} are trained on 16 NVIDIA Tesla V100 32G GPUs with a total batch size of 480K input frames for 400K update steps. While \our{}-\textit{limited} is trained with a batch size of 80K input frames for 400K steps. We optimize the models using AdamW optimizer, warming up the learning rate to a peak of 5e-4 over the first 32K updates, followed by a linear decay. We set \(\beta = 0.5\) for the spectrogram flux loss and \(\gamma = 1.0\) for the stop prediction loss. For the KL divergence loss, we set \(\lambda = 0\) for the first 10K steps to ensure stable training, and \(\lambda = 0.1\) thereafter.

% \vspace{-1mm}
\subsection{Evaluation Settings}

Following recent works \cite{wang2023valle, chen2024valle2}, we use LibriSpeech test-clean set~and screen audios with lengths ranging from 4 to 10 seconds for zero-shot evaluation.
% We use LibriSpeech test-clean set for zero-shot TTS evaluation, ensuring that none of the speakers from this corpus are included in the training data. Following recent works \cite{wang2023valle, chen2024valle2}, we screen audios with the length ranging from 4 to 10 seconds for evaluation.
We assess \our{} under two inference schemes: (1) \textit{Continuation}: We use the text transcript and the first 3 seconds of the audio as the prompt, expecting the model to seamlessly synthesize the subsequent portion of the speech; (2) \textit{Cross-sentence}: Using a reference utterance and its transcript as the prompt, and given the text of a target utterance, expecting the model to generate the corresponding speech while retaining the characteristics of the reference speaker.

To assess the naturalness, robustness, and speaker similarity of \our{}, we
employ multiple subjective and objective metrics:

\input{tables/table1.objective}

% \vspace{-2mm}
\paragraph{WER}
% Existing codec language models often face robustness issues, including word deletion, insertion, replacement, and predicting endless silence or noise. 
To assess robustness and intelligibility, we perform ASR on synthesized speech using both a Conformer-Transducer model\footnote{\url{https://huggingface.co/nvidia/stt_en_conformer_transducer_xlarge}} \cite{gulati20conformer} and HuBERT-Large ASR model\footnote{\url{https://huggingface.co/facebook/hubert-large-ls960-ft}} \cite{hsu2021hubert}. We calculate WER between the transcripts and the ground truth text. We use WER$_C$ and WER$_H$ to denote WER obtained from the two ASR systems.

% \vspace{-2mm}
\paragraph{SIM}
Speaker similarity reflects the in-context learning capability of zero-shot TTS models. We utilize WavLM-TDNN\footnote{\url{https://github.com/microsoft/UniSpeech/tree/main/downstreams/speaker_verification}} \cite{chen2022wavlm} to extract speaker embedding vectors from the original speech prompt and the generated speech. The cosine distance between them is then calculated to measure speaker similarity, denoted as SIM.
% with the range of [-1, 1]. 
% We compute SIM$_r$ and SIM$_o$, where SIM$_r$ measures the similarity between the synthesized speech and the speech prompt reconstructed from the mel-spectrogram with the vocoder, whereas SIM$_o$ measures the similarity with respect to the original speech prompt. Considering that SIM$_r$ is not comparable among systems using different acoustic tokens, we recommend referring to SIM$_o$. 

% \vspace{-2mm}
\paragraph{Subjective metrics} 
% We engaged native English speakers to participate as contributors in a crowd-sourced evaluation. 
% Each utterance was assessed by at least 10 native speakers from various perspectives. The crowd-sourcing platform also oversaw and validated the testing process and results.
Three mean opinion scores (MOS) are assessed: (1) MOS for assessing speech quality; (2) Similarity MOS (SMOS) for measuring speaker similarity between the speech prompt and the generated speech; and (3) Comparative MOS (CMOS) for evaluating the comparative naturalness of the synthesized speech against ground truth. The assessment criteria is detailed in Appendix \ref{appendix:mos}.
% For the MOS and SMOS evaluations, each test sample is rated on a scale from 1 to 5, in 0.5-point increments. Higher scores indicate more positive evaluations. For the CMOS evaluation,  the ground-truth sample and the generated sample are presented in random order to participants, who assign scores from -3 (much worse than the baseline) to 3 (much better than the baseline), with intervals of 1. We evaluate 40 samples from our test set, with one sample per speaker. 

% \paragraph{Inference time} 
% % \our{} supports adjusting the reduction factor $r$ to predict multiple mel-spectrogram frames at each time step, thereby shortening the sequence length and accelerating both model training and inference. 
% We evaluate the efficiency of \our{} with different reduction factors by measuring the inference time for generating 10 s speech segments, averaged over ten trials, and compared it with other systems.

%% file: tables/table1.objective.tex
\begin{table*}[t]

  % \footnotesize
  \small

  \centering 
\vspace{-1mm}
  \setlength{\tabcolsep}{2.5mm}
  \begin{tabular}{l c ccc ccc}

    \toprule
 \multirow{2}{*}{\textbf{System}} & \multirow{2}{*}{\textbf{\makecell{Training Data\\Hours}}} & \multicolumn{3}{c}{\textbf{Continuation}} & \multicolumn{3}{c}{\textbf{Cross-Sentence}} \\
 \cmidrule(lr){3-5} \cmidrule(l){6-8}
    &  & WER$_C$  & WER$_H$  & SIM   & WER$_C$  & WER$_H$  & SIM  \\
  \midrule
   Ground Truth & - & 1.61  & 2.15 & 0.668  & 1.61  & 2.15 & 0.779\\
    Ground Truth (mel-spectrogram) & - &1.64  & 2.24 & 0.617 & 1.64  & 2.24 & 0.732 \\
    Ground Truth (EnCodec, 8 codebooks) & - &1.65 & 2.33 & 0.593 & 1.65 & 2.33 & 0.710 \\

    \midrule
    RALL-E \citep{xin2024ralle} & 44K &-&-&-& 2.5  & 2.8 &0.49   \\
    ELLA-V  \citep{song2024ellav} * & 960 & 2.10 & 2.91 & 0.303  & 7.15 & 8.90 & 0.307 \\
    VALL-E R \citep{han2024valler}  \dag & 960 & 1.58 & 2.32 & 0.363 & 3.18 & 3.97 & 0.365 \\
    CLaM-TTS \citep{kim2024clamtts} & 55K & -  & 2.36 & 0.477 & -  & 5.11 & 0.495 \\
    VALL-E \citep{wang2023valle} & 60K & -  &  3.8 & 0.508 & -  & 5.9 & 0.580 \\

   VALL-E 2 \citep{chen2024valle2} \dag & 50K & 1.6 & 2.32 & 0.504 & {1.5} & 2.44 &  {0.643} \\
 
   Voicebox \citep{le2024voicebox} & 60K & -  & {2.0} & \textbf{0.593} & - & \textbf{1.9} & \textbf{0.662} \\
   \midrule                 
   \our{} &50K& {1.47} & \textbf{1.98} & {0.508}  & \textbf{1.47} & {2.10} & 0.625 \\
  \our{}-\textit{R2} &50K& \textbf{1.45} & 2.02 & 0.489 &   {1.50} & 2.14 & 0.608 \\
  \our{}-\textit{R3}&50K& 1.52 & 2.10 & 0.462 & 1.51 & 2.19 & 0.570 \\
  \our{}-\textit{R4} &50K& 1.59 & 2.10 & 0.437 & 1.56 & 2.30 & 0.532 \\
  \our{}-\textit{R5} &50K& 1.66 & 2.25 & 0.410 & 1.96 & 2.72 & 0.506   \\
   % \our{}-\textit{R8}  \\
  \our{}-\textit{limited} & 960 & 1.53 & 2.22 & 0.480 & 2.21 & 2.80 & 0.591\\

\bottomrule
\end{tabular}
% \vspace{-2mm}
  \caption{Objective performance comparison on \textit{continuation} and \textit{cross-sentence} zero-shot speech synthesis tasks. \our{}-\textit{Rx} denotes the model is with a reduction factor of \textit{x}. \our{}-\textit{limited} denotes the model is trained on smaller-scale corpus. *We quote \citet{han2024valler}'s reproduction results, which demonstrate better performance. 
  \dag We evaluate metrics not reported in the original paper, using the audios provided by the authors.}

  \label{tab:obj_1}
% \vspace{-4mm}

\end{table*}

%% file: 05.results_and_discussion.tex
% \vspace{-1mm}
\section{Results and Discussion}

% \vspace{-1mm}

% \footnote{We do not include the results of NaturalSpeech 3 reported in \citet{ju2024naturalspeech3} in our comparisons, as it uses a 40-utterance test set which is different with other studies.}

In this section, we compare the speech synthesis performance of \our{} with various systems,
% \footnote{We do not include the results of NaturalSpeech 3 \cite{ju2024naturalspeech3} in the comparisons, as it uses a different 40-sample test set.}
and discuss ablation study and inference efficiency.
Particularly, we would like to point out that, as shown in Table \ref{tab:obj_1}, the ground-truth speech reconstructed from mel-spectrograms demonstrates better robustness and speaker similarity compared to the speech reconstructed from EnCodec codes. This confirms the hypothesis that discrete codec codes, originally designed for audio compression, sacrifice fidelity compared to the continuous mel-spectrogram.

%%%%%%%%%%%%%%%%%%%%%%%%%%%%%%%%%%%%%%%%

% \vspace{-1mm}
\subsection{Objective Evaluation}
% \vspace{-0.5mm}
As illustrated in Table \ref{tab:obj_1}, the proposed \our{} outperforms VALL-E and all its variants on the \textit{continuation} zero-shot speech synthesis task, and is comparable to \mbox{VALL-E 2} on the \textit{cross-sentence} task. Most importantly, it presents a much more concise and efficient paradigm for audio language modeling without vector quantization.

\our{} significantly outperforms VALL-E in both robustness and speaker similarity, achieving a 47.9\% relative reduction in WER$_H$ on continuation task and a 64.4\% reduction on cross-sentence task.
ELLA-V and VALL-E R explicitly introduce monotonic alignment mechanisms to improve robustness, as reflected in the WERs. However, it comes at the cost of a significant decrease in SIM. CLaM-TTS demonstrated acceptable performance on continuation task, but its performance is limited on cross-sentence task. 
It introduces more complex assumptions and therefore an intricate structure. 
Despite both being single-pass models, 
\our{} outperforms by a large margin featuring a simpler topology.
% 待定
VALL-E 2 uses repetition-aware sampling and employs Vocos \cite{siuzdak2024vocos} as its codec decoder, demonstrating results on par with ours. For continuation task, \our{} reveals better robustness and speaker similarity. This indicates that \our{} exhibits superior zero-shot capabilities with even shorter prompts, highlighting its in-context learning ability. We attribute this advantage to our direct prediction of spectrograms, which encompass richer acoustic cues compared to discrete codes. 
For cross-sentence task, although \our{} falls slightly behind in the objective SIM metric, it still significantly surpasses VALL-E 2 in subjective metrics, as evidenced in Table \ref{tab:mos}. 
We attribute the slight difference in this objective metric to the bias of the speaker verification model, considering that \our{} achieves a higher SIM compared to VALL-E 2 (0.680 vs. 0.662), when evaluate using
another well-recognized speaker verification model, 
ECAPA-TDNN.

Although Voicebox shows better SIM than \our{}, this gap can be partially attributed to their proprietary vocoder, which was trained on a 60K-hour corpus. In contrast, \our{} utilizes an open-source vocoder trained on the 585-hour LibriTTS. Moreover, Voicebox requires both duration prediction and phoneme tokens for synthesis, whereas \our{} only requires BPE text tokens.

% 分析不同reduction factor的效果
Referring to previous mel-spectrograms prediction works, \our{} can accelerate training and inference by predicting multiple frames through an adjustable reduction factor $r$. We observe that as $r$ increases, robustness remains consistently high for both continuation and cross-sentence tasks. Although SIM declines due to the prediction of multiple frames at once, \our{} still remarkably outperforms most recent works in both WER and SIM, as shown in Table \ref{tab:obj_1}. \our{}-\textit{limited}, trained on the smaller-scale LibriSpeech corpus, also demonstrates superior performance compared to VALL-E and its variants, except for VALL-E 2.

\input{tables/table2.objective_5times}

A potential use of \our{} is to set a larger $r$ while sampling multiple times, selecting the candidate with the highest SIM to the prompt as the final output. This strategy enhances performance while reducing inference time, as the process can be executed in parallel on the GPU. 
To explore the upper bound performance of \our{} with different $r$, we report five-time sampling results in Table \ref{tab:obj_5}. In this setup, we sample five times for each test utterance and select the candidate with the best score for each metric. \our{}s consistently exhibit high robustness across different $r$ settings, yielding much lower WER than ground truth.

% 使用ecapa tdnn，melle比valle2的sim效果更好，sim: 0.680 vs 0.0.662

% 很有意思melle和valle2的WER$_C$是comparable，但valle2的WER$_H$相对低。考虑到Comformer-Transducer是一个更先进(SOTA)的ASR模型，我们可以认为valle2的鲁棒性/稳定性实际上较差于melle

%%%%%%%%%%%%%%%%%%%%%%%%%%%%%%%%%%%%%%%%
% \vspace{-2mm}
\subsection{Subjective Evaluation}
% \vspace{-0.5mm}

We conducted subjective evaluations using a crowd-source human rating system to assess 
MOS, SMOS, and CMOS, which correspond to overall speech quality, speaker similarity, and naturalness of the synthesized speech, respectively. 
We evaluated 40 samples from the test set, selecting one sample per speaker. Each speaker's previous utterance from the official test set list was used as a prompt to synthesize the target speech audio.
We use the original 16 kHz audios as the ground truth in the evaluations, unlike VALL-E 2 paper which utilizes 24 kHz upsampled audios as the ground truth.

\input{tables/table3.mos}

As shown in Table \ref{tab:mos}, \our{}'s synthesized speech is more favorably received by human listeners, achieving the best performance across all metrics compared to other systems. Remarkably, \our{} attains an SMOS score even higher than the ground truth (4.40 vs. 3.94), highlighting its exceptional capability to capture and retain the speaker's characteristics.
Furthermore, \our{} achieves speech quality on par with human-level (CMOS: -0.032 vs. 0, with $p\text{-value} > 0.1$ according to a t-test), indicating that \our{} can generate accurate and highly natural speech.
Besides, \our{}-\textit{R2}, despite sacrificing some performance for efficiency, still outperforms VALL-E 2 in MOS and SMOS. 

Additionally, we found that \our{}'s latent sampling, which avoids manually designed sampling strategy for discrete codec codes, enables it to generate more stable and natural speech compared to both \mbox{VALL-E 2} and VALL-E. 
We recommend visiting our demo website for more information.

\input{tables/table5.ablation}

\subsection{Ablation Study}
% \vspace{-0.5mm}

To assess the effectiveness of the proposed methods, we conduct a series of ablation studies on \our{}.
% , reporting both single-time and five-time sampling results. 
If the latent sampling is marked as disabled in Table \ref{tab:ablation}, it will degrade into a simple linear layer without reparameterization.

As illustrated in Table \ref{tab:ablation}, both the proposed latent sampling method and the spectrogram flux loss significantly enhance the robustness and speaker similarity of the synthesized speech. The improvements are particularly pronounced in cross-sentence task, suggesting that the proposed methods substantially facilitate longer sequence modeling. The phenomenon is also evident in the five-time sampling setup, as shown in Appendix \ref{appendix:ablation}.
We also conduct an experiment where latent sampling is enabled during training but disabled during inference. The results indicate that latent sampling during inference leads to more robust and natural outputs.

We would like to emphasize the role of latent sampling in improving speaker similarity. Compared to spectrogram flux loss, latent sampling offers relatively less improvement in WER, yet it provides comparable gains in SIM. This suggests that the primary function of latent sampling is to capture and preserve the speaker characteristics present in the speech prompt. On the other hand, spectrogram flux loss improves SIM partly by enhancing \our{}'s robustness and ensuring the accurate generation of semantic context.

%%%%%%%%%%%%%%%%%%%%%%%%%%%%%%%%%%%%%%%%

% \vspace{-1mm}
\subsection{Efficiency Comparison}
% \vspace{-0.5mm}
% % \subsection{Inference Efficiency versus Performance}
%

We compare the inference time for generating 10-second speech segments across different models. Since VALL-E and VALL-E 2 (without code grouping) share the identical architecture, their inference time can be considered the same. As shown in \mbox{Table \ref{tab:efficiency}}, \our{} is more efficient than VALL-E 2, as it forgoes the NAR inference steps, thereby reducing both computational and spatial complexity. 
% Moreover, the mel-spectrograms are extracted at approximately 62 Hz, whereas the EnCodec codes used by VALL-E is at 75 Hz.
% \mbox{ELLA-V} predicts text phoneme tokens along with acoustic tokens, implementing a monotonic alignment mechanism to drive robust outputs. This can introduce considerable overhead in both inference and training.
% , while RALL-E predicts duration and pitch tokens for each phoneme before generating the acoustic tokens.  
% Both approaches incorporate additional supervision and conditions to drive robust outputs, but also introduce considerable overhead in both inference and training.
By setting the reduction factor \( r \), the training and inference processes of \our{} can be accelerated by approximately \( r \) times --  \our{}-\textit{R2} halves the inference time, while \our{}-\textit{R4} reduces it to one quarter, surpassing VALL-E R, CLaM-TTS, and Voicebox. Despite predicting multiple frames per step, they still demonstrate satisfactory performance, as revealed in Table \ref{tab:obj_1} and Table \ref{tab:obj_5}.

% \input{tables/table4.efficiency}
\input{tables/table4.efficiency}

%% file: tables/table2.objective_5times.tex
\begin{table}[t]

  % \footnotesize
  \small

  \centering 
  \setlength{\tabcolsep}{0.3mm}
  % \renewcommand{\arraystretch}{0.95}
  % \vspace{-1mm}
  \begin{tabular}{ l ccc ccc}

    \toprule
 \multirow{2}{*}{\textbf{System}} & \multicolumn{3}{c}{\textbf{Continuation}} & \multicolumn{3}{c}{\textbf{Cross-Sentence}} \\
 \cmidrule(r){2-4} \cmidrule(l){5-7}
     & WER$_C$  & WER$_H$  & SIM   & WER$_C$  & WER$_H$ & SIM   \\

 \midrule

   Ground Truth & 1.61  & 2.15  & 0.668  & 1.61  & 2.15 & 0.779\\
   \midrule
   \our{}   & \textbf{1.03} & 1.49 &  \textbf{0.561}  & \textbf{0.70} & \textbf{1.07} &  \textbf{0.663}  \\
   \our{}-\textit{R2}  & 1.04 & \textbf{1.47} & 0.542 & 0.77 & 1.12 & 0.647          \\
   \our{}-\textit{R3} & 1.12 & 1.54 & 0.512 & 0.86 & 1.17 & 0.608 \\
   \our{}-\textit{R4} & 1.11 & 1.52 & 0.487 & 0.76 & 1.08 & 0.571 \\
   \our{}-\textit{R5} & 1.05 & 1.52 & 0.463 & 0.93 & 1.38 & 0.547  \\

 \our{}-\textit{limited} & 1.04 & 1.57 & 0.533 &1.04 & 1.50  & 0.631 \\

\bottomrule
\end{tabular}

% \vspace{-2mm}

 \caption{Comparison of five-time sampling performance with different reduction factors. The results indicate the upper bound of the systems' performance. }

  \label{tab:obj_5}
% \vspace{-2mm}

\end{table}

%% file: tables/table3.mos.tex
\begin{table}[t]

    \small
  \centering 
  \setlength{\tabcolsep}{2.3mm}
  % \renewcommand{\arraystretch}{0.98}

    % \vspace{-5mm}
  
  \begin{tabular}{ l ccc}

    \toprule
 \textbf{System} 
   & MOS & SMOS  & CMOS   \\

 \midrule
 
    Ground Truth  & 4.29$_{\pm 0.16}$ & 3.94$_{\pm 0.25}$ & 0.000 \\
    \midrule
    YourTTS \citeyearpar{casanova2022yourtts} & 2.41$_{\pm 0.24}$ & 2.62$_{\pm 0.25}$ & -2.162 \\
    VALL-E \citeyearpar{wang2023valle} & 3.18$_{\pm 0.23}$ & 3.50$_{\pm 0.25}$ & -0.912      \\
    VALL-E 2 \citeyearpar{chen2024valle2}  & 4.08$_{\pm 0.18}$ & 3.88$_{\pm 0.25}$ & -0.085\\
    \midrule
    \our{}  & \textbf{4.20}$_{\pm 0.20}$ & \textbf{4.40}$_{\pm 0.22}$ & \textbf{-0.032}  \\
    \our{}-\textit{R2}  & 4.14$_{\pm 0.19}$ & 4.18$_{\pm 0.24}$ & -0.252 \\
    
 % \our{}-\textit{limited} \\

\bottomrule
\end{tabular}

% \vspace{-2mm}
  \caption{Subjective evaluation under cross-sentence task for 40 samples from LibriSpeech test-clean set. }

  \label{tab:mos}

    % \vspace{-5mm}
\end{table}

%% file: tables/table5.ablation.tex
\begin{table}[t]
  \small

  \centering 
  \setlength{\tabcolsep}{1.mm}
  % \renewcommand{\arraystretch}{0.95}

% \vspace{-5mm}
  
  \begin{tabular}{cc ccc ccc }

    \toprule
   \multirow{2}{*}{\textbf{\makecell[c]{LS}}}   & \multirow{2}{*}{\textbf{\makecell[c]{SFL}}}  & \multicolumn{3}{c}{\textbf{Continuation}} & \multicolumn{3}{c}{\textbf{Cross-Sentence}} \\
 \cmidrule(lr){3-5} \cmidrule(l){6-8}
  &   & WER$_C$  & WER$_H$   & SIM   & WER$_C$  & WER$_H$ & SIM  \\
  \midrule
   % \multicolumn{10}{l}{\textit{Five-Time Sampling}} \\ 
  % \midrule
  \xmark    & \xmark   & 6.41 & 6.91  &  0.483  & 23.21 & 23.65  & 0.518 \\

     \cmark    & \xmark  & 3.57 & 4.07 & 0.486 &  10.36  & 10.87 & 0.584          \\
     \xmark    &  \cmark  &  2.03& 2.61& 0.506 & 5.31 & 5.90  & 0.602  \\ 
    \omark    & \cmark & 1.54& 2.13 &  0.506 & 2.10 & 2.72  & 0.615 \\
     \cmark    & \cmark  & \textbf{1.47} & \textbf{1.98} &  \textbf{0.508}  & \textbf{1.47} & \textbf{2.10}  & \textbf{0.625}    \\
  
\bottomrule
\end{tabular}

% \vspace{-2mm}
  \caption{Ablation study on the latent sampling (\textbf{LS}) and the spectrogram flux loss (\textbf{SFL}). 
  The \omark denotes that latent sampling is enabled only during training.}

  \label{tab:ablation}
% \vspace{-2mm}
\end{table}

%% file: tables/table4.efficiency.tex
% 这个表格直接用的hanbing的结果

\begin{table}[t]

  \centering
  \small

    \setlength{\tabcolsep}{3.mm}
  \begin{tabular}{lcccc}
    \toprule
    \textbf{System}   &\textbf{AR Steps}  & \textbf{Infer. Time (s)} \\
    \midrule
    % AudioLM~\citeyearpar{borsos2023audiolm} & 750 * 8  &  $>$ 40 & -  &  $>$ 40 \\ 
    % VALL-E~\citeyearpar{wang2023valle} *   & 750  &  10.27    \\
    % ELLA-V~\citeyearpar{song2024ellav} *   & \url{~}105 * 2 + 750   & 15.76   \\
    % RALL-E~\citeyearpar{xin2024ralle} * & $\sim$ 105 + 750  & 12.28 \\
    VALL-E R \citeyearpar{han2024valler} *     & 375   &  3.67 \\

    VoiceBox~\citeyearpar{le2024voicebox} \dag & -  & 6.4 (64 NFE) \\
    CLaM-TTS~\citeyearpar{kim2024clamtts} \dag &  - & 4.15 \\
    \midrule
    VALL-E~\citeyearpar{wang2023valle}   & 750  &  7.32    \\
    VALL-E 2~\citeyearpar{chen2024valle2}   & 750  &  7.32    \\
    \midrule
    \our{}  & 625 &   5.49 \\
    \our{}-\textit{R2}  & 312 &  2.76 \\
    \our{}-\textit{R4}  & 156  &  1.40 \\

    \bottomrule
  \end{tabular}
    % \vspace{-2mm}
  \caption{Inference time for generating 10-second speech segments. 
  % The "105" refers to the number of auxiliary tokens the model need to predict. 
  *Quoted from \citet{han2024valler}; 
  \dag Quoted from \citet{kim2024clamtts}.}
  \label{tab:efficiency}
% \vspace{-3mm}
\end{table}

%% file: 06.conclusion.tex
% \vspace{-1mm}
\section{Conclusion}
% \vspace{-1mm}
We present a continuous-valued token based language modeling approach for zero-shot text-to-speech synthesis, thereby eliminating the use of vector quantization.
By exploring the potential of mel-spectrograms within the paradigm of language modeling, the proposed \our{} directly predicts continuous-valued tokens conditioned on text content and speech prompt. This approach obviates the need for the two-stage training and inference procedures typical of neural codec language models like VALL-E, and can further accelerate decoding by setting the reduction factor. 
With the aid of latent sampling and spectrogram flux loss, \our{} is capable of producing more diverse and robust predictions, attaining highly natural speech comparable to human performance in subjective evaluations.
% \footnote{In addition to the model described in this paper, we also trained a \our{} model for \textbf{Mandarin} text-to-speech, using the same model configurations and training settings as the English version. Please check our demo page.}

% \input{tables/table4.efficiency}

%% file: 07.misc.tex
% Acknowledgements

% Broader Impacts

\section*{Limitations}
% \label{appendix:broader}
% \paragraph{Limitations} 
Despite \our{}'s promising performance and concise topology, we acknowledge several limitations. 
First, the quality of synthesized speech can be limited by the ability of the vocoder utilized.  We anticipate performance improvements by training a more powerful vocoder on a large-scale corpus, as demonstrated by Voicebox \cite{le2024voicebox}. 
Second, we conduct evaluation on English-only LibriSpeech test set. The Multi-lingual setting like VALL-E X \cite{zhang2023vallex} on various dataset will be explored in our future work. 
Third, we adopt only the mel-spectrogram as the target continuous acoustic representation. Future research will explore other continuous representations, such as VAE latent hidden states.

\section*{Broader Impacts and Ethical Statements}
We envision advancing the development of speech synthesis by distilling the methodology of audio language modeling to its fundamental principles, eliminating the complexity of heavy codebooks. 
The proposed approach can substantially reduce the training and inference costs of large-scale audio generation models while improving performance.

\our{} is purely a research project. 
% Currently, we have no plans to incorporate \our{} into a product or expand access to the public. 
\our{} could synthesize speech that maintains speaker identity and could be used for education, entertainment, journalistic, self-authored content, accessibility features, interactive voice response systems, translation, chatbot, and so on. While \our{} can speak in a voice like a voice talent, the similarity and naturalness of the generated speech depend on the length and quality of the speech prompt, the background noise, as well as other factors. It may carry potential risks in the misuse of the model, such as spoofing voice identification or impersonating a specific speaker. We conducted the experiments under the assumption that the user agrees to be the target speaker in speech synthesis. If the model is generalized to unseen speakers in the real world, it should include a protocol to ensure that the speaker approves the use of their voice and a synthesized speech detection model.

All data and pre-trained models used are publicly available and are used under following licenses: Creative Commons BY 4.0 License, Creative Commons CC0 License, Creative Commons BY-NC-ND 4.0 License, Creative Commons BY-SA 4.0 License, MIT license, and Apache-2.0 license.

%%%%% If you suspect that \our{} is being used in a manner that is abusive or illegal or infringes on your rights or the rights of other people, you can report it at the Report Abuse Portal.

\section*{Acknowledgments}
This work is partially supported by the CUHK MoE-Microsoft Key Laboratory of Human-Centric Computing and Interface Technologies, and a grant from the HKSARG Research Grants Council's Theme-based Research Grant Scheme (Project No. T45-407/19N).

%% file: 98.appendix.tex
\clearpage
\newpage

\setcounter{secnumdepth}{2}
\onecolumn
\appendix
\section{Appendix}

\subsection{Derivation of Kullback-Leibler (KL) Divergence Loss}
\label{appendix:derivation}
We assume that $\bm{z}_t$ follows a multivariate Gaussian distribution where each dimension is independent. Combining equation (\ref{eq:gaussian}), the KL divergence loss among $T$ time steps can be analytically computed as
{
\begin{align}
&\mathcal{L}_\text{KL}(\bm{y}, \bm{z}) = \sum_{t=0}^{T-1} D_{\text{KL}}(p_{\bm{\theta}}(\bm{z}_t \mid \bm{e}_t) \parallel p(\bm{z}_t))  \notag\\
&= \sum_{t=0}^{T-1} D_{\text{KL}}\left(\mathcal{N}(\bm{\mu}_t, \mathrm{diag}(\bm{\sigma}_t^2)) \parallel  \mathcal{N}(\bm{y}_t, \bm{I}) \right)  \notag\\
% &=  \sum_{t=0}^{T-1} \int p_\theta(\bm{z}_t \mid \bm{e}_t) \log \frac{p_\theta(\bm{z}_t \mid \bm{e}_t)}{p(\bm{z}_t)} \, d\bm{z}_t  \notag  \\
&=  \sum_{t=0}^{T-1} \sum_{i=1}^d \int \frac{1}{\sqrt{2\pi \bm{\sigma}_t^2[i]}} e^{-\frac{(x - \bm{\mu}_t[i])^2}{2 \bm{\sigma}_t^2[i]}} \log \frac{e^{-(x - \bm{\mu}_t[i])^2/2\bm{\sigma}^2_t[i]} / \sqrt{2\pi\bm{\sigma}^2_t[i]}}{e^{-(x-\bm{y}_t[i])^2/2}/\sqrt{2\pi}} \, dx  \notag  \\
&=  \sum_{t=0}^{T-1} \sum_{i=1}^d \int \frac{1}{\sqrt{2\pi \bm{\sigma}_t^2[i]}} e^{-\frac{(x - \bm{\mu}_t[i])^2}{2 \bm{\sigma}_t^2[i]}} \log \frac{e^{\left(\left(x-\bm{y}_t[i]\right)^2-(x - \bm{\mu}_t[i])^2/\bm{\sigma}_t^2[i]\right)/2}}{\sqrt{\bm{\sigma}_t^2[i]}} \, dx \notag\\
&=\frac{1}{2} \sum_{t=0}^{T-1} \sum_{i=1}^d \int \frac{1}{\sqrt{2\pi \bm{\sigma}_t^2[i]}} e^{-\frac{(x - \bm{\mu}_t[i])^2}{2 \bm{\sigma}_t^2[i]}} 
\left( (x-\bm{y}_t[i])^2 - (x -\bm{\mu}_t[i])^2/\bm{\sigma}_t^2[i]  -\log \bm{\sigma}_t^2[i]  \right) \, dx \notag \\
&=\frac{1}{2} \sum_{t=0}^{T-1} \sum_{i=1}^d \left( \int \frac{1}{\sqrt{2\pi \bm{\sigma}_t^2[i]}} e^{-\frac{(x - \bm{\mu}_t[i])^2}{2 \bm{\sigma}_t^2[i]}} \left( \left( x - \bm{\mu}_t[i] \right) + \left( \bm{\mu}_t[i] - \bm{y}_t[i] \right) \right)^2 \, dx \right) - 1 - \log \bm{\sigma}_t^2[i]\notag \\
&=\frac{1}{2} \sum_{t=0}^{T-1} \sum_{i=1}^d \left( \bm{\sigma}_t^2[i] + (\bm{\mu}_t[i] - \bm{y}_t[i])^2 - 1 - \log \bm{\sigma}_t^2[i] \right)
\notag\\
&=\frac{1}{2} \sum_{t=0}^{T-1} ( \|\bm{\sigma}_t\|^2_2 + \|\bm{\mu}_t - \bm{y}_t\|^2_2 - d - \sum_{i=1}^d \log \bm{\sigma}_t^2[i] )
\end{align} 
}
where \( d \) is the dimensionality of the feature space.

\subsection{Mel-Spectrogram Extraction Protocol}
\label{appendix:mel}
We extract log-magnitude mel spectrograms from resampled 16 kHz audios as the target continuous speech representation throughout this work. To extract mel-spectrograms, we apply a 1024-point short-time Fourier transform (STFT) using the Hann window function, with a window length of 1024 and a hop length of 256. We then apply an 80-dimensional mel-filter with the frequency range of \mbox{80 Hz} to 7600 Hz. Finally, we take the base-10 logarithm of the resulting output as the final representation.

\subsection{Training Details}
\label{appendix:train}
\our{} are trained on 16 NVIDIA Tesla V100 32G GPUs with a total batch size of 480K input frames for 400K update steps. While \our{}-\textit{limited} is trained with a batch size of 80K input frames for 400K steps. We optimize the models using AdamW optimizer, warming up the learning rate to a peak of 5e-4 over the first 32K updates, followed by a linear decay. We set \(\beta = 0.5\) for the spectrogram flux loss and \(\gamma = 1.0\) for the stop prediction loss. For the KL divergence loss, we set \(\lambda = 0\) for the first 10K steps to ensure stable training, and \(\lambda = 0.1\) thereafter.

\subsection{Detailed Subjective Assessment Criteria}
\label{appendix:mos}
% The proposed method is
% evaluated using three types of mean opinion scores (MOS): (1) MOS for assessing speech quality;
% (2) Similarity MOS (SMOS) for measuring speaker similarity between the speech prompt and the
% generated speech; and (3) Comparative MOS (CMOS) for evaluating the comparative naturalness of
% the synthesized speech against the original ground truth audio. 
% We evaluate 40 samples from our test set, with one sample for each speaker.

% For the MOS and SMOS evaluations, the test audio samples are sent to a crowd-source human rating system where each sample is rated by at least 10 native English speakers on a scale from 1 to 5, with 0.5-point increments. The higher the score, the more positive the evaluation results.
% For the CMOS evaluation, each pair of utterances is presented in random order to ten native English speakers. Participants are asked to assign a score ranging from -3 to 3, with intervals of 1. A score of -3 indicates that the new system is much worse than the baseline, while a score of 3 indicates that the new system is much better.

We engaged native English speakers with experience in speech annotation and evaluation to participate as contributors in a crowd-sourced evaluation. The crowd-sourcing platform also oversaw and validated the testing process and results.

We evaluate 40 samples from our test set, with one sample for each speaker. Each utterance was assessed by at least 10 contributors from various perspectives. 
Three types of mean opinion scores (MOS) are assessed: (1) MOS for assessing speech quality; (2) Similarity MOS (SMOS) for measuring speaker similarity between the speech prompt and the generated speech; and (3) Comparative MOS (CMOS) for evaluating the comparative naturalness of the synthesized speech against the original ground truth audio. 
For MOS and SMOS evaluations, each test sample is rated on a scale from 1 to 5, in increments of 0.5 points. Higher scores indicate more positive evaluations. For the CMOS evaluation, the ground truth sample and the generated sample are presented in random order to the participants, who assign scores from -3 (much worse than the baseline) to 3 (much better than the baseline), with intervals of 1. 

\subsection{Ablation Study with Five-Time Sampling}
\label{appendix:ablation}
To further demonstrate the effectiveness of the proposed method, we also report the results of the ablation study with five-time sampling. In this setup, we sampled five times for each test utterance and selected the candidate with the best score for each metric for reporting. The upper half of Table \ref{tab:appendix_ablation} presents the results for single-time sampling, which is same as Table \ref{tab:ablation} in the main text. The lower half shows the results for five-time sampling.

As shown in Table \ref{tab:appendix_ablation}, the proposed latent sampling method and the spectrogram flux loss significantly enhance the robustness and speaker similarity of the synthesized speech.
This improvement is evident in both single-time sampling and five-time sampling setups.

\input{tables/tableA1.ablaion}

%% file: tables/tableA1.ablaion.tex
\begin{table*}[h]

  \small

  \centering 
\setcounter{table}{0}
\renewcommand{\thetable}{A\arabic{table}}
  
  \begin{tabular}{ccc ccc ccc }

    \toprule
 &  \multirow{2}{*}{\textbf{\makecell[c]{Latent\\Sampling}}}   & \multirow{2}{*}{\textbf{\makecell[c]{Spectrogram\\Flux Loss}}}  & \multicolumn{3}{c}{\textbf{Continuation}} & \multicolumn{3}{c}{\textbf{Cross-Sentence}} \\
 \cmidrule(r){4-6} \cmidrule(l){7-9}
  &  &   & WER$_C$  & WER$_H$   & SIM   & WER$_C$  & WER$_H$ & SIM  \\
  \midrule
   % \multicolumn{10}{l}{\textit{Five-Time Sampling}} \\ 
  % \midrule
 \multirow{5}{*}{\makecell[c]{Single-Time\\Sampling}} & \xmark    & \xmark   & 6.41 & 6.91  &  0.483  & 23.21 & 23.65  & 0.518 \\

    & \cmark    & \xmark  & 3.57 & 4.07 & 0.486 &  10.36  & 10.87 & 0.584          \\
   &  \xmark    &  \cmark  &  2.03& 2.61& 0.506 & 5.31 & 5.90  & 0.602  \\ 
   & \omark    & \cmark & 1.54& 2.13 &  0.506 & 2.10 & 2.72  & 0.615 \\
  &   \cmark    & \cmark  & \textbf{1.47} & \textbf{1.98} &  \textbf{0.508}  & \textbf{1.47} & \textbf{2.10}  & \textbf{0.625}    \\
  
     \midrule
 \multirow{5}{*}{\makecell[c]{Five-Time\\Sampling}} & \xmark    & \xmark   & 3.74 & 4.15 &  0.536  & 17.69 & 18.00  & 0.569 \\

    & \cmark    & \xmark  & 1.18 & 1.63  & 0.546 &  2.41  & 2.86  & 0.641          \\
   &  \xmark    &  \cmark & 1.17 & 1.65  & 0.551 & 1.74 & 2.13  & 0.644  \\ 
   &   \omark    & \cmark & 1.10 & 1.50  & 0.552 & 1.07 & 1.47& 0.645\\ 
  &   \cmark    & \cmark  &  \textbf{1.03} & \textbf{1.49} &   \textbf{0.561}  & \textbf{0.70} & \textbf{1.07} & \textbf{0.663}    \\

\bottomrule
\end{tabular}

% \vspq

  \caption{Ablation study on the effectiveness of latent sampling and the spectrogram flux loss. 
  The \omark denotes that latent sampling is enabled during training but disabled during inference.}

  \label{tab:appendix_ablation}

\end{table*}

%% file: 00.main.bbl
\begin{thebibliography}{42}
\providecommand{\natexlab}[1]{#1}

\bibitem[{Anastassiou et~al.(2024)Anastassiou, Chen, Chen, Chen, Chen, Chen, Cong, Deng, Ding, Gao et~al.}]{anastassiou2024seed}
Philip Anastassiou, Jiawei Chen, Jitong Chen, Yuanzhe Chen, Zhuo Chen, Ziyi Chen, Jian Cong, Lelai Deng, Chuang Ding, Lu~Gao, et~al. 2024.
\newblock {Seed-TTS}: A family of high-quality versatile speech generation models.
\newblock \emph{arXiv preprint arXiv:2406.02430}.

\bibitem[{Ao et~al.(2022)Ao, Wang, Zhou, Wang, Ren, Wu, Liu, Ko, Li, Zhang, Wei, Qian, Li, and Wei}]{ao2021speecht5}
Junyi Ao, Rui Wang, Long Zhou, Chengyi Wang, Shuo Ren, Yu~Wu, Shujie Liu, Tom Ko, Qing Li, Yu~Zhang, Zhihua Wei, Yao Qian, Jinyu Li, and Furu Wei. 2022.
\newblock {S}peech{T}5: Unified-modal encoder-decoder pre-training for spoken language processing.
\newblock In \emph{Proceedings of the 60th Annual Meeting of the Association for Computational Linguistics}, pages 5723--5738.

\bibitem[{Bai et~al.(2024)Bai, Likhomanenko, Zhang, Gu, Aldeneh, and Jaitly}]{bai2024dmel}
He~Bai, Tatiana Likhomanenko, Ruixiang Zhang, Zijin Gu, Zakaria Aldeneh, and Navdeep Jaitly. 2024.
\newblock {dMel}: Speech tokenization made simple.
\newblock \emph{arXiv preprint arXiv:2407.15835}.

\bibitem[{Betker(2023)}]{betker2023tortoise}
James Betker. 2023.
\newblock Better speech synthesis through scaling.
\newblock \emph{arXiv preprint arXiv:2305.07243}.

\bibitem[{Borsos et~al.(2023)Borsos, Sharifi, Vincent, Kharitonov, Zeghidour, and Tagliasacchi}]{borsos2023soundstorm}
Zal{\'a}n Borsos, Matt Sharifi, Damien Vincent, Eugene Kharitonov, Neil Zeghidour, and Marco Tagliasacchi. 2023.
\newblock {SoundStorm}: Efficient parallel audio generation.
\newblock \emph{arXiv preprint arXiv:2305.09636}.

\bibitem[{Brown et~al.(2020)Brown, Mann, Ryder, Subbiah, Kaplan, Dhariwal, Neelakantan, Shyam, Sastry, Askell et~al.}]{brown2020language}
Tom Brown, Benjamin Mann, Nick Ryder, Melanie Subbiah, Jared~D Kaplan, Prafulla Dhariwal, Arvind Neelakantan, Pranav Shyam, Girish Sastry, Amanda Askell, et~al. 2020.
\newblock Language models are few-shot learners.
\newblock \emph{Advances in neural information processing systems}, 33:1877--1901.

\bibitem[{Casanova et~al.(2022)Casanova, Weber, Shulby, Junior, G{\"o}lge, and Ponti}]{casanova2022yourtts}
Edresson Casanova, Julian Weber, Christopher~D Shulby, Arnaldo~Candido Junior, Eren G{\"o}lge, and Moacir~A Ponti. 2022.
\newblock {YourTTS}: Towards zero-shot multi-speaker {TTS} and zero-shot voice conversion for everyone.
\newblock In \emph{International Conference on Machine Learning}, pages 2709--2720.

\bibitem[{Chen et~al.(2024{\natexlab{a}})Chen, Wang, Ren, Li, Zhao, Li, Cai, Guo, Zhang, Xiong et~al.}]{chen2024ntp}
Liang Chen, Zekun Wang, Shuhuai Ren, Lei Li, Haozhe Zhao, Yunshui Li, Zefan Cai, Hongcheng Guo, Lei Zhang, Yizhe Xiong, et~al. 2024{\natexlab{a}}.
\newblock Next token prediction towards multimodal intelligence: A comprehensive survey.
\newblock \emph{arXiv preprint arXiv:2412.18619}.

\bibitem[{Chen et~al.(2024{\natexlab{b}})Chen, Liu, Zhou, Liu, Tan, Li, Zhao, Qian, and Wei}]{chen2024valle2}
Sanyuan Chen, Shujie Liu, Long Zhou, Yanqing Liu, Xu~Tan, Jinyu Li, Sheng Zhao, Yao Qian, and Furu Wei. 2024{\natexlab{b}}.
\newblock {VALL-E} 2: Neural codec language models are human parity zero-shot text to speech synthesizers.
\newblock \emph{arXiv preprint arXiv:2406.05370}.

\bibitem[{Chen et~al.(2022)Chen, Wang, Chen, Wu, Liu, Chen, Li, Kanda, Yoshioka, Xiao et~al.}]{chen2022wavlm}
Sanyuan Chen, Chengyi Wang, Zhengyang Chen, Yu~Wu, Shujie Liu, Zhuo Chen, Jinyu Li, Naoyuki Kanda, Takuya Yoshioka, Xiong Xiao, et~al. 2022.
\newblock {WavLM}: Large-scale self-supervised pre-training for full stack speech processing.
\newblock \emph{IEEE Journal of Selected Topics in Signal Processing}, 16(6):1505--1518.

\bibitem[{D{\'e}fossez et~al.(2023)D{\'e}fossez, Copet, Synnaeve, and Adi}]{dfossez2023encodec}
Alexandre D{\'e}fossez, Jade Copet, Gabriel Synnaeve, and Yossi Adi. 2023.
\newblock High fidelity neural audio compression.
\newblock \emph{Transactions on Machine Learning Research}.
\newblock Featured Certification, Reproducibility Certification.

\bibitem[{Du et~al.(2024)Du, Chen, Zhang, Hu, Lu, Yang, Hu, Zheng, Gu, Ma et~al.}]{du2024cosyvoice}
Zhihao Du, Qian Chen, Shiliang Zhang, Kai Hu, Heng Lu, Yexin Yang, Hangrui Hu, Siqi Zheng, Yue Gu, Ziyang Ma, et~al. 2024.
\newblock {CosyVoice}: A scalable multilingual zero-shot text-to-speech synthesizer based on supervised semantic tokens.
\newblock \emph{arXiv preprint arXiv:2407.05407}.

\bibitem[{Eskimez et~al.(2024)Eskimez, Wang, Thakker, Li, Tsai, Xiao, Yang, Zhu, Tang, Tan et~al.}]{eskimez2024e2}
Sefik~Emre Eskimez, Xiaofei Wang, Manthan Thakker, Canrun Li, Chung-Hsien Tsai, Zhen Xiao, Hemin Yang, Zirun Zhu, Min Tang, Xu~Tan, et~al. 2024.
\newblock E2 {TTS}: Embarrassingly easy fully non-autoregressive zero-shot {TTS}.
\newblock \emph{arXiv preprint arXiv:2406.18009}.

\bibitem[{Gulati et~al.(2020)Gulati, Qin, Chiu, Parmar, Zhang, Yu, Han, Wang, Zhang, Wu, and Pang}]{gulati20conformer}
Anmol Gulati, James Qin, Chung-Cheng Chiu, Niki Parmar, Yu~Zhang, Jiahui Yu, Wei Han, Shibo Wang, Zhengdong Zhang, Yonghui Wu, and Ruoming Pang. 2020.
\newblock Conformer: Convolution-augmented transformer for speech recognition.
\newblock In \emph{Proc. Interspeech 2020}, pages 5036--5040.

\bibitem[{Han et~al.(2024)Han, Zhou, Liu, Chen, Meng, Qian, Liu, Zhao, Li, and Wei}]{han2024valler}
Bing Han, Long Zhou, Shujie Liu, Sanyuan Chen, Lingwei Meng, Yanming Qian, Yanqing Liu, Sheng Zhao, Jinyu Li, and Furu Wei. 2024.
\newblock {VALL-E R}: Robust and efficient zero-shot text-to-speech synthesis via monotonic alignment.
\newblock \emph{arXiv preprint arXiv:2406.07855}.

\bibitem[{Hsu et~al.(2021)Hsu, Bolte, Tsai, Lakhotia, Salakhutdinov, and Mohamed}]{hsu2021hubert}
Wei-Ning Hsu, Benjamin Bolte, Yao-Hung~Hubert Tsai, Kushal Lakhotia, Ruslan Salakhutdinov, and Abdelrahman Mohamed. 2021.
\newblock {HuBERT}: Self-supervised speech representation learning by masked prediction of hidden units.
\newblock \emph{IEEE/ACM Trans. Audio, Speech and Lang. Proc.}, 29:3451–3460.

\bibitem[{Jiang et~al.(2023)Jiang, Ren, Ye, Liu, Zhang, Yang, Ji, Huang, Wang, Yin et~al.}]{jiang2023mega}
Ziyue Jiang, Yi~Ren, Zhenhui Ye, Jinglin Liu, Chen Zhang, Qian Yang, Shengpeng Ji, Rongjie Huang, Chunfeng Wang, Xiang Yin, et~al. 2023.
\newblock {Mega-TTS}: Zero-shot text-to-speech at scale with intrinsic inductive bias.
\newblock \emph{arXiv preprint arXiv:2306.03509}.

\bibitem[{Ju et~al.(2024)Ju, Wang, Shen, Tan, Xin, Yang, Liu, Leng, Song, Tang et~al.}]{ju2024naturalspeech3}
Zeqian Ju, Yuancheng Wang, Kai Shen, Xu~Tan, Detai Xin, Dongchao Yang, Yanqing Liu, Yichong Leng, Kaitao Song, Siliang Tang, et~al. 2024.
\newblock {Naturalspeech} 3: Zero-shot speech synthesis with factorized codec and diffusion models.
\newblock \emph{arXiv preprint arXiv:2403.03100}.

\bibitem[{Kang et~al.(2024)Kang, Yang, Yao, Kuang, Yang, Guo, Lin, and Povey}]{kang2024libriheavy}
Wei Kang, Xiaoyu Yang, Zengwei Yao, Fangjun Kuang, Yifan Yang, Liyong Guo, Long Lin, and Daniel Povey. 2024.
\newblock Libriheavy: a 50,000 hours {ASR} corpus with punctuation casing and context.
\newblock In \emph{ICASSP 2024-2024 IEEE International Conference on Acoustics, Speech and Signal Processing (ICASSP)}, pages 10991--10995.

\bibitem[{Kim et~al.(2024)Kim, Lee, Chung, and Cho}]{kim2024clamtts}
Jaehyeon Kim, Keon Lee, Seungjun Chung, and Jaewoong Cho. 2024.
\newblock {CLaM-TTS}: Improving neural codec language model for zero-shot text-to-speech.
\newblock In \emph{The Twelfth International Conference on Learning Representations}.

\bibitem[{Kingma and Welling(2014)}]{kingma2014vae}
Diederik~P Kingma and Max Welling. 2014.
\newblock Auto-encoding variational bayes.
\newblock In \emph{The International Conference on Learning Representations}.

\bibitem[{Kong et~al.(2020)Kong, Kim, and Bae}]{kong2020hifigan}
Jungil Kong, Jaehyeon Kim, and Jaekyoung Bae. 2020.
\newblock {HiFi-GAN}: Generative adversarial networks for efficient and high fidelity speech synthesis.
\newblock In \emph{Advances in Neural Information Processing Systems}, volume~33, pages 17022--17033.

\bibitem[{Kumar et~al.(2023)Kumar, Seetharaman, Luebs, Kumar, and Kumar}]{kumar2023dac}
Rithesh Kumar, Prem Seetharaman, Alejandro Luebs, Ishaan Kumar, and Kundan Kumar. 2023.
\newblock High-fidelity audio compression with improved {RVQGAN}.
\newblock \emph{Advances in Neural Information Processing Systems}, 36:27980--27993.

\bibitem[{{\L}ajszczak et~al.(2024){\L}ajszczak, C{\'a}mbara, Li, Beyhan, van Korlaar, Yang, Joly, Mart{\'\i}n-Cortinas, Abbas, Michalski et~al.}]{lajszczak2024base}
Mateusz {\L}ajszczak, Guillermo C{\'a}mbara, Yang Li, Fatih Beyhan, Arent van Korlaar, Fan Yang, Arnaud Joly, {\'A}lvaro Mart{\'\i}n-Cortinas, Ammar Abbas, Adam Michalski, et~al. 2024.
\newblock {BASE TTS}: Lessons from building a billion-parameter text-to-speech model on {100K} hours of data.
\newblock \emph{arXiv preprint arXiv:2402.08093}.

\bibitem[{Le et~al.(2023)Le, Vyas, Shi, Karrer, Sari, Moritz, Williamson, Manohar, Adi, Mahadeokar, and Hsu}]{le2024voicebox}
Matthew Le, Apoorv Vyas, Bowen Shi, Brian Karrer, Leda Sari, Rashel Moritz, Mary Williamson, Vimal Manohar, Yossi Adi, Jay Mahadeokar, and Wei-Ning Hsu. 2023.
\newblock Voicebox: Text-guided multilingual universal speech generation at scale.
\newblock In \emph{Thirty-seventh Conference on Neural Information Processing Systems}.

\bibitem[{Li et~al.(2019)Li, Liu, Liu, Zhao, and Liu}]{li2019transformertts}
Naihan Li, Shujie Liu, Yanqing Liu, Sheng Zhao, and Ming Liu. 2019.
\newblock Neural speech synthesis with transformer network.
\newblock \emph{Proceedings of the AAAI Conference on Artificial Intelligence}, page 6706–6713.

\bibitem[{Li et~al.(2024{\natexlab{a}})Li, Tian, Li, Deng, and He}]{kaiming2024autoregressive}
Tianhong Li, Yonglong Tian, He~Li, Mingyang Deng, and Kaiming He. 2024{\natexlab{a}}.
\newblock Autoregressive image generation without vector quantization.
\newblock \emph{arXiv preprint arXiv:2406.11838}.

\bibitem[{Li et~al.(2024{\natexlab{b}})Li, Han, Raghavan, Mischler, and Mesgarani}]{li2024styletts}
Yinghao~Aaron Li, Cong Han, Vinay Raghavan, Gavin Mischler, and Nima Mesgarani. 2024{\natexlab{b}}.
\newblock {StyleTTS} 2: Towards human-level text-to-speech through style diffusion and adversarial training with large speech language models.
\newblock \emph{Advances in Neural Information Processing Systems}, 36.

\bibitem[{Liu et~al.(2024)Liu, Wang, Inoue, Bai, and Li}]{liu2024ardit}
Zhijun Liu, Shuai Wang, Sho Inoue, Qibing Bai, and Haizhou Li. 2024.
\newblock Autoregressive diffusion transformer for text-to-speech synthesis.
\newblock \emph{arXiv preprint arXiv:2406.05551}.

\bibitem[{OpenAI(2023)}]{OpenAI2023GPT4TR}
OpenAI. 2023.
\newblock {GPT-4} technical report.
\newblock \emph{arXiv preprint arXiv:2303.08774}.

\bibitem[{Panayotov et~al.(2015)Panayotov, Chen, Povey, and Khudanpur}]{panayotov2015librispeech}
Vassil Panayotov, Guoguo Chen, Daniel Povey, and Sanjeev Khudanpur. 2015.
\newblock {LibriSpeech}: An {ASR} corpus based on public domain audio books.
\newblock In \emph{ICASSP}, pages 5206--5210.

\bibitem[{Puvvada et~al.(2024)Puvvada, Rao~Koluguri, Dhawan, Balam, and Ginsburg}]{puvvada2024discrete}
Krishna~C. Puvvada, Nithin Rao~Koluguri, Kunal Dhawan, Jagadeesh Balam, and Boris Ginsburg. 2024.
\newblock Discrete audio representation as an alternative to mel-spectrograms for speaker and speech recognition.
\newblock In \emph{ICASSP 2024 - 2024 IEEE International Conference on Acoustics, Speech and Signal Processing (ICASSP)}, pages 12111--12115.

\bibitem[{Ren et~al.(2019)Ren, Ruan, Tan, Qin, Zhao, Zhao, and Liu}]{ren2019fastspeech}
Yi~Ren, Yangjun Ruan, Xu~Tan, Tao Qin, Sheng Zhao, Zhou Zhao, and Tie{-}Yan Liu. 2019.
\newblock {FastSpeech}: Fast, robust and controllable text to speech.
\newblock In \emph{NeurIPS}, pages 3165--3174.

\bibitem[{Siuzdak(2024)}]{siuzdak2024vocos}
Hubert Siuzdak. 2024.
\newblock Vocos: Closing the gap between time-domain and {Fourier}-based neural vocoders for high-quality audio synthesis.
\newblock In \emph{The Twelfth International Conference on Learning Representations}.

\bibitem[{Song et~al.(2024)Song, Chen, Wang, Ma, and Chen}]{song2024ellav}
Yakun Song, Zhuo Chen, Xiaofei Wang, Ziyang Ma, and Xie Chen. 2024.
\newblock {ELLA-V}: Stable neural codec language modeling with alignment-guided sequence reordering.
\newblock \emph{arXiv preprint arXiv:2401.07333}.

\bibitem[{Tschannen et~al.(2023)Tschannen, Eastwood, and Mentzer}]{tschannen2023givt}
Michael Tschannen, Cian Eastwood, and Fabian Mentzer. 2023.
\newblock {GIVT}: Generative infinite-vocabulary transformers.
\newblock \emph{arXiv preprint arXiv:2312.02116}.

\bibitem[{Vyas et~al.(2023)Vyas, Shi, Le, Tjandra, Wu, Guo, Zhang, Zhang, Adkins, Ngan et~al.}]{vyas2023audiobox}
Apoorv Vyas, Bowen Shi, Matthew Le, Andros Tjandra, Yi-Chiao Wu, Baishan Guo, Jiemin Zhang, Xinyue Zhang, Robert Adkins, William Ngan, et~al. 2023.
\newblock Audiobox: Unified audio generation with natural language prompts.
\newblock \emph{arXiv preprint arXiv:2312.15821}.

\bibitem[{Wang et~al.(2023)Wang, Chen, Wu, Zhang, Zhou, Liu, Chen, Liu, Wang, Li et~al.}]{wang2023valle}
Chengyi Wang, Sanyuan Chen, Yu~Wu, Ziqiang Zhang, Long Zhou, Shujie Liu, Zhuo Chen, Yanqing Liu, Huaming Wang, Jinyu Li, et~al. 2023.
\newblock Neural codec language models are zero-shot text to speech synthesizers.
\newblock \emph{arXiv preprint arXiv:2301.02111}.

\bibitem[{Wang et~al.(2017)Wang, Skerry-Ryan, Stanton, Wu, Weiss, Jaitly, Yang, Xiao, Chen, Bengio, Le, Agiomyrgiannakis, Clark, and Saurous}]{wang2017tacotron}
Yuxuan Wang, R.J. Skerry-Ryan, Daisy Stanton, Yonghui Wu, Ron~J. Weiss, Navdeep Jaitly, Zongheng Yang, Ying Xiao, Zhifeng Chen, Samy Bengio, Quoc Le, Yannis Agiomyrgiannakis, Rob Clark, and Rif~A. Saurous. 2017.
\newblock Tacotron: Towards end-to-end speech synthesis.
\newblock In \emph{Proc. Interspeech 2017}, pages 4006--4010.

\bibitem[{Xin et~al.(2024)Xin, Tan, Shen, Ju, Yang, Wang, Takamichi, Saruwatari, Liu, Li et~al.}]{xin2024ralle}
Detai Xin, Xu~Tan, Kai Shen, Zeqian Ju, Dongchao Yang, Yuancheng Wang, Shinnosuke Takamichi, Hiroshi Saruwatari, Shujie Liu, Jinyu Li, et~al. 2024.
\newblock {RALL-E}: Robust codec language modeling with chain-of-thought prompting for text-to-speech synthesis.
\newblock \emph{arXiv preprint arXiv:2404.03204}.

\bibitem[{Zeghidour et~al.(2021)Zeghidour, Luebs, Omran, Skoglund, and Tagliasacchi}]{zeghidour2021soundstream}
Neil Zeghidour, Alejandro Luebs, Ahmed Omran, Jan Skoglund, and Marco Tagliasacchi. 2021.
\newblock {SoundStream}: An end-to-end neural audio codec.
\newblock \emph{IEEE/ACM Transactions on Audio, Speech, and Language Processing}, 30:495--507.

\bibitem[{Zhang et~al.(2023)Zhang, Zhou, Wang, Chen, Wu, Liu, Chen, Liu, Wang, Li et~al.}]{zhang2023vallex}
Ziqiang Zhang, Long Zhou, Chengyi Wang, Sanyuan Chen, Yu~Wu, Shujie Liu, Zhuo Chen, Yanqing Liu, Huaming Wang, Jinyu Li, et~al. 2023.
\newblock Speak foreign languages with your own voice: Cross-lingual neural codec language modeling.
\newblock \emph{arXiv preprint arXiv:2303.03926}.

\end{thebibliography}
